\documentclass[manuscript,screen]{acmart}

\usepackage{multirow}
\usepackage{subcaption}
\usepackage{hyperref}

\AtBeginDocument{%
  \providecommand\BibTeX{{%
    \normalfont B\kern-0.5em{\scshape i\kern-0.25em b}\kern-0.8em\TeX}}}

\setcopyright{acmcopyright}
\copyrightyear{2024}
\acmYear{2024}
\acmDOI{XXXXXXX.XXXXXXX}

\begin{document}

\title{Trustworthy Distributed AI Systems: Robustness, Privacy, and Governance}

\author{Wenqi Wei}
\affiliation{%
  \institution{Fordham University}
  \streetaddress{113 West 60th Street}
  \city{New York City, NY}
  \country{USA}}
\email{wenqiwei@fordham.edu}


\author{Ling Liu}
\affiliation{%
  \institution{Georgia Institute of Technology}
  \streetaddress{North Avenue}
  \city{Atlanta, GA}
  \country{USA}}
\email{ling.liu@cc.gatech.edu}

\renewcommand{\shortauthors}{Wenqi Wei and Ling Liu}

\begin{abstract}
Emerging Distributed AI systems are revolutionizing big data computing and data processing capabilities with growing economic and societal impact. However, recent studies have identified new attack surfaces and risks caused by security, privacy, and fairness issues in AI systems. 
In this paper, we review representative techniques, algorithms, and theoretical foundations for trustworthy distributed AI through robustness guarantee, privacy protection, and fairness awareness in distributed learning. We first provide a brief overview of alternative architectures for distributed learning, discuss inherent vulnerabilities for security, privacy, and fairness of AI algorithms in distributed learning, and analyze why these problems are present in distributed learning regardless of specific architectures. Then we provide a unique taxonomy of countermeasures for trustworthy distributed AI, covering (1) robustness to evasion attacks and irregular queries at inference, and robustness to poisoning attacks, Byzantine attacks, and irregular data distribution during training; (2) privacy protection during distributed learning and model inference at deployment; and (3) AI fairness and governance with respect to both data and models.
We conclude with a discussion on open challenges and future research directions toward trustworthy distributed AI, such as the need for trustworthy AI policy guidelines, the AI responsibility-utility co-design, and incentives and compliance. 
\end{abstract}



\maketitle


\section{Introduction}

Distributed Artificial Intelligence (AI) is a class of distributed learning algorithms and optimizations that span from swarm intelligence to multi-agent technologies. Compared to centralized machine learning (ML) algorithms, distributed learning offers a unique property $-$ the default privacy: Instead of the centralized training data, distributed learning allows a population of geographically distributed edge clients to jointly train a global ML model while keeping their sensitive or proprietary data local. At each learning round, the participating clients only need to share their local training model updates with the aggregation server~\cite{mcmahan2017communication}. 
On the one hand, distributed AI provides countless new opportunities for collaborative learning of a global model from a large population of geographically distributed participants without centralizing the training data. On the other hand, distributed AI systems also open doors to new attack surfaces and vulnerabilities concerning security, privacy, and fairness. Distributed learning is an open system,  consisting of geographically distributed participating clients (subscribers) from multiple administrative domains. Therefore, these clients are more exposed to different types of compromises and disruptive events, including biases in data collection.  

To develop effective safeguards for trustworthy distributed AI, it is essential to understand the requirements of trustworthy AI in the context of distributed learning systems. This includes examining the known vulnerabilities due to irregular data,  security problems, privacy violations, and fairness issues.   
First, 
training data for distributed learning is disparate across multiple clients and are non-IID in nature~\cite{zhao2018federated}. As a result, trustworthy distributed learning requires strong robustness against irregular data, such as imbalanced data, long-tailed data, and out-of-distribution data. 
Second, 
trustworthy distributed learning needs robust solutions to mitigate data corruption at compromised clients. Example corruptions include data poisoning attacks~\cite{tolpegin2020data,chow2023stdlens} and data thefts due to the presence of semi-honest but curious agents at either local clients or the aggregation server (e.g., gradient leakage attack~\cite{wei2021gradient,wei2021gradient_tifs}). 
Third, 
the aggregation server for distributed learning may also be compromised due to Byzantine attacks or model poisoning attacks by throttling the usability of the global model or changing the optimization goal of distributed learning. 
Furthermore, a well-trained global model from distributed learning can also be compromised during model inference (deployment) by evasion attacks. Evasion attacks maliciously manipulate the input data or the prediction output of the model to make the model prediction fail miserably. 
At the data input level, the evasion attack is designed by corrupting the query input data using an adversarial example generated by adding a small amount of strategically learned noise to the benign query example~\cite{goodfellow2014explaining,carlini2017towards}. Although the small adversarial noise used to corrupt the input data is imperceptible to humans, adversarial examples can effectively deceive the well-trained model to make targeted mistakes or random prediction errors. 
At the output level, adversaries may launch reconstruction inference on prediction output through query probing, which may cause unauthorized partial or full disclosure of private training data using attribute/property inference attack~\cite{melis2019exploiting} or membership inference attack~\cite{shokri2017membership,truex2019demystifying}. Model inversion attack~\cite{fredrikson2015model} is another type of output inference attack through query probing, aiming to infer the decision boundary, resulting in leaking sensitive attributes about the query examples, such as sensitive health data, when such queries are from individual patients. 
Finally,
biases in data collection may result from the data collection and/or the data preprocessing process, such as sampling errors and annotation biases due to known human biases, ranging from confirmation bias, over-generation bias to over-confidence bias. Biases in data tend to propagate differently due to different loss optimizations used during model training. The impact of biases in data may amplify the biases in model prediction (biases in algorithms).

\subsection{Requirement for Trustworthy Distributed AI} 
Motivated by the above requirement analysis, we argue that robustness, privacy, and fairness are three critical pillars for ensuring and strengthening the trustworthiness of distributed AI. 

\textbf{AI robustness} is the ability of AI systems to cope with errors during model learning or inference. 
A model is considered robust if its prediction output is consistently accurate, even if one or more input variables or assumptions are changed due to unforeseen circumstances. Robustness of a model should be tested against changes to input, for example, by adding some noise to the test data and altering the magnitude of the noise. Strong robustness indicates that the model will perform well with new data and different sources of noise.   

\textbf{AI privacy} is the ability of AI systems to cope with sensitive data leakages during model learning or deployment.
A model is considered privacy-preserving if the model is considered safe from privacy leakages in training and inference. During model training, unauthorized access to intermediate model update parameters should not leak information about the private training data involved. Similarly, during model deployment, unauthorized inference on its prediction output should be prevented against sensitive information leakage about the private data used in training the model or against the potential linkage to a private individual based on the sensitive data (e.g., disease diagnoses) in the prediction output. As a result, the 
privacy of a model should be tested against different inference methods that attempt to use the prediction output of the model to reconstruct the sensitive training data or to link user identity with the sensitive attributes of the query input. 

\textbf{AI fairness} is the ability of AI systems to cope with biases in data and biases in algorithms. Bias is a preference or prejudice against an individual or a particular group. Fairness is a socially defined concept and the subjective practice of using AI without favoritism or discrimination in societal and social context~\cite{fiske1998stereotyping,mehrabi2021survey}. Given a set of training data, a family of bias metrics is used to describe how a model can perform/predict differently for distinct groups within the data. It is widely recognized that human bias comes in many forms, e.g., data collection and data annotation can bring about biases in data. 
Fairness of a model should be tested against different sources of data biases and algorithmic biases, especially when the training data can be partitioned in multiple different ways. Each partitioning will result in various groups. It is important to make sure that the predictions of a model are calibrated for each group, in order to alleviate or avoid systemically overestimating or underestimating the probability of the outcome for one of the groups. High-level fairness assurance indicates that the model has been re-examined and refined by identifying and eradicating different sources of bias and tested by fairness defined by pre-established ethical principles~\cite{mehrabi2021survey}. 
With the increasing number of regulations on AI robustness, AI privacy, and AI fairness in recent years, it becomes even more challenging to incorporate diverse governance policies in the design of trustworthy distributed AI systems and applications. 

\subsection{Contributions and Scopes}
In this paper, we review representative techniques and theoretical foundations for trustworthy distributed AI through robustness guarantee, privacy protection, and fairness awareness in distributed learning.  
The survey is organized as follows: Based on a brief overview of alternative architectures for distributed learning, we first outline some representative attacks that threaten the robustness, privacy, and fairness of AI algorithms in distributed learning, regardless of the specific architectures in use (Section~\ref{sec2}). Then we provide a comprehensive review of robustness countermeasures to mitigate evasion attacks and irregular queries at inference, as well as poisoning attacks, Byzantine attacks, and irregular data distribution during federated learning (Section~\ref{sec3}). Next, we review the representative privacy-enhancing techniques to enable privacy-preserving distributed learning against different types of unauthorized data leakages (Section~\ref{sec4}). We dedicate Section~\ref{sec5} to briefly overview privacy leakage and deception problems, such as adversarial manipulation of rewards, in the distributed multi-armed bandit learning systems, and discuss the representative countermeasures.
Finally, we examine different sources of bias and important fairness-enhancing techniques. We also describe the critical roles of (i) data governance with respect to privacy-preserving and fairness-aware data collection and (ii) model governance with respect to incorporating and enforcing analytical and operational constraints on data input and prediction output of a model (Section~\ref{sec6}). We conclude the paper with a discussion on some open challenges and future research directions toward trustworthy distributed AI, including the need for (i) governance policy guidelines, (ii) responsibility-utility co-design, and (iii) incentive and compliance.

\begin{figure}[t]
\centerline{\includegraphics[scale=.66]{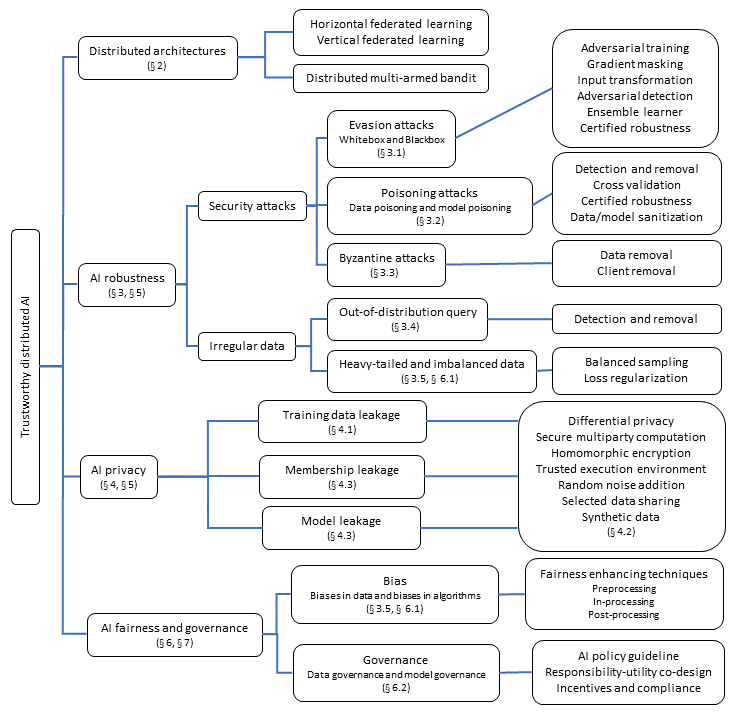}}
\vspace{-0.3cm}
\caption{\small Taxonomy of surveyed approaches in trustworthy distributed AI.} 
\label{fig:metaphor}
\vspace{-0.4cm}
\end{figure}

To the best of our knowledge, this is the first survey that reviews the robustness, privacy, fairness, and governance of AI models in distributed learning. 
This paper can benefit AI researchers and AI software developers with an in-depth understanding of the importance of trustworthy distributed AI and the collection of responsible AI techniques for ensuring robustness, privacy, and fairness against input, output, and computation corruption in the context of distributed learning. At the same time, the survey may also help cybersecurity researchers, engineers, and practitioners, who apply AI technology to solving information security problems, to gain advanced knowledge about the importance of trustworthy AI and the potential issues of AI models. These issues range from privacy leakage, and data bias induced unfairness to the drastic degradation of model robustness due to data poisoning and input deception. 
We argue that trustworthy distributed AI systems by design should have built-in trust governance, trust-guard structures, and robust countermeasures against adversarial and systemic disruptions. 
To guide readers with the taxonomy of surveyed approaches in trustworthy distributed AI throughout the paper, we 
create a visualization metaphor in Figure~\ref{fig:metaphor}. Our taxonomy serves as a structured roadmap for organizing the diverse set of vulnerabilities and countermeasures in the field of trustworthy distributed AI. It helps researchers and practitioners navigate the complexities of building systems that are robust, privacy-preserving, and fair while also highlighting the interconnections between these aspects, leading to a more holistic understanding of trustworthiness in distributed AI.

\section{Reference Framework: An Overview}

\label{sec2}

This section gives a brief overview of the reference framework we use to organize this paper. We first categorize existing distributed learning systems into two broad categories based on the network structures used to design and optimize across multiple participants. Then we briefly review a set of known vulnerabilities, followed by the countermeasures for enhancing the robustness, privacy, and fairness in the context of trustworthy distributed learning. 


\subsection{Representative Architectures for Distributed Learning}
\label{sec2.1}


Federated learning and distributed multi-armed bandits are the two dominating architectures in distributed AI literature. 
In federated learning~\cite{yang2019federated}, the aggregation server will initiate the training of a global model and progress in a pre-configured number of rounds (convergence condition). In each round of distributed learning, only a small percentage of the clients is selected to participate in this round of joint training. First, the server distributes the global model to each contributing client. Next, each of the contributing clients will perform local training over its local data and send the local model update parameters to the aggregation server. Finally, the server aggregates the local model update to produce the updated global model and distributes it for the next round of distributed learning~\cite{mcmahan2017communication}. Most off-the-shelf machine learning frameworks provide software modules to deploy distributed training tasks on edge devices, e.g., the TensorFlow Lite module~\cite{abadi2016tensorflow} and PyTorch Mobile module~\cite{paszke2019pytorch}. 

\subsubsection{Horizontal and Vertical Federated Learning} 
Federated learning literature is further categorized into horizontal and vertical federated learning based on the sample space and the feature space of the training data. The sample space refers to the total training data collection distributed across the total $N$ clients. The feature space refers to the set of attributes/properties used to describe each sample in the sample space. 

Horizontal federated learning~\cite{yang2019federated} assumes that all participating clients can use the same global model to perform the local model training. Training data from each client is a subset of samples in the input space, and each is described with the full set of features from the feature space. One can view the distributed collection of training data from all $N$ clients as a horizontal partition of the whole sample space of training data, similar to the row partitioning in a distributed relational database\footnote{\url{https://publications.opengroup.org/standards/data-mgmt}}. 
In horizontal federated learning, the aggregation of local model updates from multiple contributing clients is relatively straightforward, with FedAVG or FedSGD~\cite{mcmahan2017communication} as the most representative method for model aggregation.  

In contrast, vertical federated learning~\cite{yang2019federated} assumes that some clients have local data with only a subset of features in the whole feature space. Other features of the same samples may reside in other clients. Hence, the feature space is vertically partitioned for some data samples. It carries some similarities to column partitioning in a distributed relational database. 
In vertical federated learning, more sophisticated aggregation methods need to be employed to combine local model updates and intermediate results at the aggregation server to jointly learn a global model. 
SplitFed~\cite{thapa2022splitfed} uses a dedicated server to orchestrate server-side weight and the main Split learning server to coordinate client-side weight. Apart from the baseline approach, 
hybrid federated learning typically combines horizontal and vertical federated learning~\cite{diao2020heterofl}. Such hybrid federated learning allows distributed learning with heterogeneous clients, especially enabling clients with limited computational capacity to participate in federated learning, even though they cannot run the full-size global model locally, providing support for enabling equitable access to diverse client communities.

\subsubsection{Distributed Multi-Armed Bandits} 
In multi-armed bandits (MAB)~\cite{bubeck2012regret}, a decision-maker faces a sequential series of decisions, each with two or more action options, coined as "arms". Each arm has an associated
probability distribution that models certain rewards for choosing the corresponding option. Upon selecting an option, the decision maker (the learner) receives a reward drawn from this option's associated probability distribution and expects to choose the option with maximized cumulative reward.
%
%
%
%
Distributed multi-agent MAB has a group of agents collaboratively improve their collective reward by communicating over a network. 
Each agent independently
makes decisions regarding their actions~\cite{liu2010distributed}. 
For example, 
\cite{martinez2019decentralized} 
studies distributed multi-armed bandits in which multiple agents face the same MAB instance. The agents collaboratively share their estimates over a fixed communication graph to design consensus-based distributed estimation algorithms to estimate the mean of rewards at each arm. 
Federated bandit is an emerging distributed online learning paradigm adopting the bandit problem in the federated setting. 
Unlike distributed multi-agent bandits, the agents only communicate with the server in certain communication rounds and cannot share their information directly. With heterogeneous reward distributions at local clients, \cite{shi2021federated} investigates efficient client-server communication and coordination protocols for federated MAB.  
%
%
%

\subsection{Trustworthy AI in Distributed Learning}

Data in Cyber systems are known to have three vulnerable points: data-at-rest, data-in-transit, and data-in-use. Numerous technologies are available to protect data-at-rest and data-in-transit in practice. For example, most federated learning systems leverage edge computing infrastructure and assume that data-at-rest and data-in-transit are encrypted. However, data-in-use resides in volatile memory (RAM), unencrypted and available to compromised applications, firmware, operating systems, and hypervisors. This leaves the data vulnerable to memory dumps and other malicious exploits before, during, and after in-memory processing. Safeguarding data-in-use remains a grand challenge because most applications need precise (unencrypted) data to compute.

 \begin{table}[t]
\centering
\caption{\small Robustness, privacy, and fairness threats covered in this paper.}
\vspace{-0.3cm}
\scalebox{0.90}{
\small{
\begin{tabular}{|cc|cc|cc|cc|c|}
\hline
\multicolumn{2}{|c|}{\multirow{2}{*}{}}                                       & \multicolumn{2}{c|}{attack target} & \multicolumn{2}{c|}{attack location} & \multicolumn{2}{c|}{attack timing}        & \multirow{2}{*}{attack effect} \\ \cline{3-8}
\multicolumn{2}{|c|}{}                                                        & \multicolumn{1}{c|}{data}  & model & \multicolumn{1}{c|}{client} & server & \multicolumn{1}{c|}{training} & inference &                                \\ \hline
\multicolumn{1}{|c|}{\multirow{2}{*}{irregular data}}  & OOD                  & \multicolumn{1}{c|}{yes}   & no    & \multicolumn{1}{c|}{no}     & yes    & \multicolumn{1}{c|}{no}       & yes       & misclassification              \\ \cline{2-9} 
\multicolumn{1}{|c|}{}                                 & imbalanced            & \multicolumn{1}{c|}{yes}   & no    & \multicolumn{1}{c|}{yes}    & no     & \multicolumn{1}{c|}{yes}      & no        & bias                           \\ \hline
\multicolumn{1}{|c|}{\multirow{4}{*}{contamination}}   & evasion              & \multicolumn{1}{c|}{yes}   & no    & \multicolumn{1}{c|}{no}     & yes    & \multicolumn{1}{c|}{no}       & yes       & misclassification              \\ \cline{2-9} 
\multicolumn{1}{|c|}{}                                 & poisoning            & \multicolumn{1}{c|}{yes}   & yes   & \multicolumn{1}{c|}{yes}    & yes    & \multicolumn{1}{c|}{yes}      & no        & misclassification              \\ \cline{2-9} 
\multicolumn{1}{|c|}{}                                 & byzantine            & \multicolumn{1}{c|}{no}    & yes   & \multicolumn{1}{c|}{yes}    & yes    & \multicolumn{1}{c|}{yes}      & no        & misclassification              \\ \cline{2-9} 
\multicolumn{1}{|c|}{}                                 & adv. bandit          & \multicolumn{1}{c|}{yes}   & no    & \multicolumn{1}{c|}{yes}    & no     & \multicolumn{1}{c|}{yes}      & no        & non-optimal regret             \\ \hline
\multicolumn{1}{|c|}{\multirow{4}{*}{privacy leakage}} & gradient leakage     & \multicolumn{1}{c|}{yes}   & no    & \multicolumn{1}{c|}{yes}    & yes    & \multicolumn{1}{c|}{yes}      & no        & data disclosure                \\ \cline{2-9} 
\multicolumn{1}{|c|}{}                                 & membership           & \multicolumn{1}{c|}{yes}   & no    & \multicolumn{1}{c|}{yes}    & yes    & \multicolumn{1}{c|}{yes}      & yes       & membership disclosure          \\ \cline{2-9} 
\multicolumn{1}{|c|}{}                                 & attributed           & \multicolumn{1}{c|}{yes}   & no    & \multicolumn{1}{c|}{yes}    & yes    & \multicolumn{1}{c|}{no}       & yes       & data disclosure                \\ \cline{2-9} 
\multicolumn{1}{|c|}{}                                 & extraction           & \multicolumn{1}{c|}{no}    & yes   & \multicolumn{1}{c|}{yes}    & yes    & \multicolumn{1}{c|}{no}       & yes       & model disclosure               \\ \hline
\multicolumn{1}{|c|}{\multirow{3}{*}{bias}}            & data collection      & \multicolumn{1}{c|}{yes}   & no    & \multicolumn{1}{c|}{yes}    & no     & \multicolumn{1}{c|}{no}       & no        & biased outcome                 \\ \cline{2-9} 
\multicolumn{1}{|c|}{}                                 & data preprocessing   & \multicolumn{1}{c|}{yes}   & no    & \multicolumn{1}{c|}{yes}    & no     & \multicolumn{1}{c|}{no}       & no        & biased outcome                 \\ \cline{2-9} 
\multicolumn{1}{|c|}{}                                 & data-driven learning & \multicolumn{1}{c|}{no}    & yes   & \multicolumn{1}{c|}{no}     & yes    & \multicolumn{1}{c|}{yes}      & no        & biased outcome                 \\ \hline
\end{tabular}
}}
\label{table:threats}
\vspace{-0.4cm}
\end{table} 

\textbf{Table~\ref{table:threats}} lists known vulnerabilities to the AI models, which may threaten the trustworthiness of distributed learning systems and applications. Most of these problems are  related to the robustness guarantee, privacy protection, and fairness awareness. 
These potential vulnerabilities for data-in-use are common to distributed learning (model training and deployment) regardless of the specific architectures used for coordinating the iterative joint learning process. First, when a client performs local model training, the training data are often unencrypted. Hence, data-in-use during local training can be compromised. Second, upon completing the local training and before sending it over the network to the aggregation server, each client will apply encryption to safeguard data-in-transit. However, the local model update prior to encryption can be compromised at client. Finally, the server needs to decrypt the local model updates shared by each client to carry out the aggregation computation. Hence, data-in-use at the aggregation server can be compromised. Similar problems happen to the rewards of arms in distributed multi-agent multi-armed bandits online learning systems. Even if we encrypt reward data-at-rest and-in-transit, the reward data-in-use is still vulnerable to all kinds of poisoning or unauthorized inference attacks.

We will organize the rest of the paper by providing an in-depth review of the sources of vulnerability under robustness, privacy, and fairness. We will review robustness countermeasures, privacy-preserving methods, and fairness-enhancing techniques. We will also describe how data governance and model governance can put humans in the loop of distributed model training and model deployment to strengthen the enforcement of trustworthy AI through explainable and responsible guidelines. 


\subsubsection{Trustworthy AI through Robustness}
Robustness of distributed AI model is the capability of the model to handle errors and disruptions in model training and model deployment in practice. We can measure the AI robustness from several dimensions regarding the safety of data-in-use, such as detection, mitigation, and prevention of misuse and abuse in both the data (input and output) and the model.  
For certain types of disruptive events, such as data imbalance-caused biases and out-of-distribution data, the best countermeasure is the capability of just-in-time detection to flag suspicious circumstances and get human experts in the loop for mitigation guidelines. For other types of disruptions, such as data contamination by evasion, poisoning, or by privacy leakages and model theft, we can broadly categorize the literature into three classes of countermeasures: (1) Auto-repair without explicit detection~\cite{madry2017towards}, (2) Auto-detection without auto-repair~\cite{xu2017feature,hendrycks2016baseline}, and (3) Auto-detect followed by auto-repair~\cite{wei2020robust}.  
We observe three scenarios for the problems that can be detected but cannot repair automatically: (i) The problems can benefit significantly by detection but are hard to repair even by human experts, such as spam or out-of-distribution data. (ii) The problems can be auto-repaired, but the detection methods lack any built-in capability for auto-repair of the errors incurred due to disruption. 
(iii) The problems are attempted for auto-repairing,  but the auto-repairing decision depends heavily on the detection methods. For example, a detection method may only flag the existence of disruptive event(s) but is unable to identify the corrupted data with high confidence. This may result in difficulty in repairing the errors incurred~\cite{wei2022thesis}.  

\subsubsection{Trustworthy AI through Privacy}
Privacy in distributed AI is the capability of AI models to protect the confidentiality and integrity of sensitive training data, such as preventing or mitigating unauthorized inference and reverse engineering of (i) intermediate computation results during training or (ii) model output and prediction result during predictive inference at model deployment. A widely recognized reason for launching reverse engineering attacks on AI models comes from the capability of model training: Upon complex model training over big data, an AI model tends to remember many intrinsic details about its training data~\cite{song2017machine}.
We will 
describe the known privacy threats in distributed learning and review representative privacy-enhancing techniques in the recent literature.
%
%
%
Privacy protection in distributed AI should safeguard the input data, the AI computation process, and the output of the model prediction, especially under the guidance and compliance of data governance and model governance frameworks and policies.

\subsubsection{Trustworthy AI through Fairness and Governance}
AI fairness is the capability of AI models to handle biases in data and algorithms. Human biases have existed for decades~\cite{fiske1998stereotyping}, including selection bias, over-generalization bias, and confirmation bias, to name a few. Furthermore, biases in data may come from data collection, such as sampling errors and data annotation errors during data preprocessing and cleaning~\cite{mehrabi2021survey}. In addition, algorithmic biases may result from analyzing the data from a limited point of view to learn the model. 
Hence, it is essential to provide fairness-aware guidelines and frameworks empowered by explainable AI methods and supported by human-in-the-loop capability. We will 
discuss fairness problems and fairness-enhancing techniques in collaboration with data and model governance.  

\section{Trustworthy Distributed AI through Robustness Guarantee}

\label{sec3}

Trustworthy Distributed AI through robustness guarantee concerns resilience of AI solutions against disruptive events during model training and deployment (edge inference). In this section, we will discuss vulnerabilities and countermeasures for enhancing and ensuring trustworthy AI through (i) robustness against evasion attacks and irregular inference queries to well-trained AI models, (ii) robustness against poisoning attacks and Byzantine attacks to distributed learning systems that jointly train an AI model from a population of geographically distributed edge clients, and (iii) robustness in the presence of imbalanced or heavy-tail training data. 


\subsection{Robustness to Evasion Attacks}

\label{sec3.1}

In this section, we first introduce evasion attacks to different machine learning tasks and then discuss a collection of representative countermeasures for improving the robustness of AI solutions against evasion attacks.

\subsubsection{Evasion Attacks during Deployment}
Evasion attack refers to adversarial examples that are maliciously crafted to misguide the well-trained AI models in generating erroneous predictions by injecting a small amount of noise into the input examples without affecting the human perception of the inputs~\cite{szegedy2013intriguing}. Evasion attack methods have reportedly led to severe damage to the trustworthiness of AI models, well trained for different types of data modalities, from image classification~\cite{szegedy2013intriguing}, object detection and video analytics~\cite{chow2020understanding}, to text classification~\cite{liang2018deep}, text understanding~\cite{li2019textbugger} and other NLP models. 
In addition to digital adversarial perturbation, recent research has shown the feasibility of physical attacks~\cite{eykholt2018robust}. Additional literature for adversarial examples can be found in the comprehensive collection\footnote{\url{https://nicholas.carlini.com/writing/2019/all-adversarial-example-papers.html}}. 

Adversarial examples are input artifacts created from benign inputs to a pre-trained target AI model at inference time by adding adversarial distortions. The goal is to cause the target model to misclassify randomly (untargeted attack) or purposefully (targeted attack) with high confidence~\cite{szegedy2013intriguing}. The attack process can be formulated as
$$min||x - x'|{|_p} \quad s.t.f(x') = {y^*},\ f(x') \ne f(x),$$
where $y^{*}$ denotes the target class label for targeted attacks or any label other than the ground-truth class for untargeted attacks~\cite{wei2020adversarial}. $p$ is the distance norm, usually in $L_0$, $L_2$, and $L_\infty$. 
Adversarial examples have witnessed a sizable body of research on various attack algorithms for the image classification task.
 \cite{szegedy2013intriguing} generates adversarial examples using an L-BFGS method to solve the general targeted problem. 
 FGSM~\cite{goodfellow2014explaining} performs a one-step gradient update along the direction of the sign of gradient at each pixel to generate adversarial examples. BIM~\cite{kurakin2016physical} extends FGSM by running the optimization for multiple iterations. JSMA~\cite{papernot2016limitations} computes the adversarial saliency map for pixel-level modification. PGD attack~\cite{madry2017towards} employs a Projected Gradient Descent to project gradients to a ball and looks at the BIM-alike iterative attack process through robust optimization.
 CW attacks~\cite{carlini2017towards} are by far the most effective attack. 
CW $L_\infty$ attack replaces the perturbation distance term with a penalty for any terms that exceed a certain threshold to prevent optimization oscillating between suboptimal solutions since $L_\infty$ norm only penalizes the largest term. 
CW $L_2$ attack converts the adversarial example to $tanh$ space to find the minimal distortion in $L_2$ distance.
CW $L_0$ attack iteratively calls CW $L_2$ attack to select the pixels with the least significant contribution to prediction and remove them from the perturbation set until the attack is successful. Deepfool~\cite{moosavi2016deepfool} is an untargeted attack that deploys an $L_2$ minimization by searching for the minimal perturbation within the polyhedron space that could fool the classifier.
The decision boundaries of nonlinear classifiers are approximated into a polyhedron by an iterative linearization
procedure. 
Due to transferability~\cite{papernot2016transferability},  
adversarial examples generated from one deep learning model can be transferred to fool other models. 

\subsubsection{Evasion Robustness by Adversarial Training} 
Adversarial training is a preventative countermeasure technique. 
Adversarial training methods train the victim model under protection by leveraging both the benign training set and adversarial examples generated using each known attack algorithm or an ensemble of them~\cite{tramer2017ensemble}. The training objective function will include the loss optimization for benign input and the loss optimization for each of the ensemble of adversarial counterparts, formally described as follows: 
$$\alpha L(x,y) + (1-\alpha)L(x_{adv},y).$$
For adversarial training with adversarial examples generated from PGD attack, \cite{madry2017towards} gives theoretical and empirical proof of local maxima's tractability and shows that PGD adversarial training significantly increases the adversarial robustness.
To address the high computational cost, \cite{shafahi2019adversarial} proposes free adversarial training by reusing the gradients computed in the backward propagation when performing a forward pass. 
A widely recognized weakness of adversarial training 
based on 
adversarial examples from known attacks 
is its limited robustness against unknown attacks~\cite{tramer2019adversarial}. 


\subsubsection{Evasion Robustness by Gradient Masking}
Gradient masking is a countermeasure technique that enhances AI robustness by modifying some components of the victim model during the training, aiming to reduce the sensitivity of the trained model to perturbations on the input data. The main idea is to obfuscate the gradient information or to use a near-zero gradient to offset or minimize the impact of gradient manipulation performed by an adversary. For example, 
\cite{gu2014towards} adds a gradient penalty term in the model training objective with a contractive autoencoder, a summation of the layer-wise Frobenius norm of the Jacobian matrix. The gradient penalty encourages the model to be invariant to small changes in directions which are irrelevant in the input space. Defensive distillation~\cite{papernot2016distillation} replaces the output layer with a modified softmax function and a temperature knot to control the extent of distillation~\cite{hinton2015distilling} after training, aiming to use the distilled model to hide gradient information of the target model from an adversary and reduce the model sensitivity to small changes in input. 
\cite{kannan2018adversarial} introduces logit pairing,  which enforces logit invariance between a clean input and its adversarial counterpart under adversarial logit pairing. The method also enforces logit invariance between any pairs of inputs under clean logit pairing. 
However, an open challenge for gradient masking methods is to determine the proper form of gradient penalty to obfuscate the gradient information with minimal impact on the accuracy performance of the trained model.


\subsubsection{Evasion Robustness by Input Transformation}
Input transformation techniques are also coined as input denoising. The main idea is to employ certain data modality-specific denoising techniques to clean the input data  before sending it as query input to the prediction model.  
The following three categories of input transformation techniques are commonly employed in image modality. The first category is to use popular image preprocessing techniques through input transformation before classification~\cite{guo2017countering}, including cropping and rescaling, bit-depth reduction, JPEG compression, Total Variance Minimization (TVM) and image quilting. Thermometer Encoding~\cite{buckman2018thermometer} quantizes and discretizes the input data and takes out the effect of small perturbations. 
The second category is learning a denoising model for input transformation by leveraging deep learning. 
Defense-GAN~\cite{samangouei2018defense} and PixelDefend~\cite{song2017pixeldefend} leverage generative models to learn the distribution of the non-adversarial dataset with the goal of projecting adversarial input to the learned non-adversarial manifold. Both methods train only on clean data in order to approximate the distribution of the data. 
The third category is to transform the input data by exploiting the neighborhood region of a given input.
Region sampling~\cite{cao2017mitigating} averages the prediction on the hypercube centered at the example.
However, input transformation methods differ by data modality and are sensitive to task-specific training datasets. A research study~\cite{he2017adversarial} shows that some input transformation defenses can be bypassed by adaptive adversaries. 

\subsubsection{Evasion Robustness by Adversarial Detection} 
Unlike the previous approaches that mitigate the adversarial examples by generating correct predictions under attack, the adversarial detection approach only detects and flags a suspicious input example. 
Feature squeezing~\cite{xu2017feature} compares the predictions between the squeezed and original inputs. The intuition is based on the observation that the input squeezing operation using different input transformation methods could remove unnecessary features from the input. 
\cite{grosse2017statistical} augments a classifier network with an additional class node representing the adversarial class. 
Magnet~\cite{meng2017magnet} trains a pair of autoencoders: detector and reformer to learn the latent space representation on training data. The method  leverages the latent patterns to detect adversarial inputs. If the input is clean, then the output of the detector looks very similar; if the input is adversarial, then the reconstruction loss between input and output will be high, and threshold detection can be performed.
However, as pointed out in~\cite{carlini2017detection}, many detection-only methods can be bypassed if the attacker is aware of the detection strategy.

\subsubsection{Evasion Robustness by Ensemble Learners} 
Ensemble robustness against evasion attack can be classified into denoising, output, and cross-layer ensembles~\cite{wei2020cross}. 
The denoising ensemble learner refers to a committee of models. Each component model uses one specific input transformation, e.g., denoising autoencoder ensemble learner~\cite{chow2019denoising}, to clean the input data before sending it to the prediction model. The denoising ensemble will generate the ensemble prediction based on the consensus method that integrates the prediction result from each of the component models, such as soft voting, majority voting, and plural voting. 
The model verification ensemble refers to a committee of models trained independently on the same task using different neural network backbone algorithms~\cite{wei2020adversarial}. Each query example will be sent in parallel to all component models of the ensemble for output verification. An important element in composing robust ensemble learners is ensemble diversity~\cite{liu2019deep,pang2019improving}. 
Specifically, $\kappa$ diversity is employed for ensemble mitigation against evasion attacks on image classification~\cite{wei2020robust}. 
Focal ensemble diversity proposed in~\cite{wu2021boosting,wu2023exploring} shows significant robustness enhancement over $\kappa$ diversity with respect to ensemble generalization performance in benign and adversarial scenarios. 
The cross-layer ensemble learners combine input denoising ensemble and output verification ensemble, providing higher robustness against different types of evasion attacks~\cite{wei2020robust}.

\subsubsection{Evasion Robustness by Certified Bound}
Research efforts on certifying the robustness of deep neural networks have been investigating certification-based verification methods, primarily for image classification problems. Certified robustness guarantees that for any given input, the classifier's prediction is constant within some bound set around the input, often defined by an L$_2$ or L$_{\infty}$ ball. 
The pioneering work~\cite{katz2017reluplex} provides a formulation of SMT solver for the ReLU activation function and uses their solver to prove/disprove local adversarial robustness for a few arbitrary combinations of input and perturbations.
Expanding over the Reluplex idea, AI2~\cite{gehr2018ai2} approximates mathematical functions with an infinite set of behaviors into logical functions which are finite and consequently computable. 
Extreme value theory~\cite{weng2018evaluating}, 
and linear programming with relaxation and duality~\cite{wong2018provable,raghunathan2018certified}  
are popular tools for certified robustness design.
Randomized Smoothing certifies the noise added to the model for adversarial robustness. \cite{cohen2019certified} proposes using a smoothed classifier to generate a tighter certification bound. \cite{lecuyer2019certified} links differential privacy~\cite{dwork2014algorithmic} with robustness against norm-bounded adversarial examples. However, most certified bound research efforts to date tend to identify specific certified bound for a given DNN structure, and rely on heuristics (e.g., bounding boxes) to compute and use such certified spatial bound~\cite{boopathy2019cnn}.

\subsubsection{Analysis of Evasion Robustness Methods} \label{sec3.1.8}

The arm race between adversaries and defense for trustworthy AI continues along all dimensions, ranging from different machine learning tasks and models to data modalities. The collection of countermeasures outlined in this section lays out some foundations and trajectories toward building the next generation of trustworthy AI systems. We argue that it is also important to study evasion robustness under black-box, grey-box, and white-box assumptions of adversaries: (i) Black-box attacks assume that attackers have no prior knowledge of the prediction defense system; (ii) Gray-box attacks assume some leakage of partial knowledge of the defense system; (iii) white-box attacks assume that attackers have full knowledge of the prediction defense system~\cite{wei2020robust}. For example, existing studies have shown that under white-box or gray-box adversaries, gradient masking is vulnerable to a suite of adaptive attacks, such as BPDA~\cite{athalye2018obfuscated}  
and EOT~\cite{athalye2018synthesizing}. 
An open challenge is to develop evasion robustness countermeasures against adaptive adversaries who strategically design attacks based on partial or complete knowledge of the defense methods.

\subsection{Robustness to Poisoning Attacks}

\label{sec3.3}

Distributed learning is typically performed over a population of geographically disparate edge clients across multiple administrative domains. Hence, 
distributed learning is inherently vulnerable to poisoning attacks from compromised participants acting as malicious insiders. 
The lack of control over participating clients could allow the malicious adversary to pretend to be a benign participant and control one or more client devices to send compromised model updates to the server. The  attackers seek to corrupt the globally trained model.

\subsubsection{Data Poisoning Attacks} 

Data poisoning attacks compromise the integrity of the training data to corrupt the global model during the iterative training process. In a distributed learning system with a population of $N$ participants (clients), data poisoning attacks may occur during local data collection and/or during local model training. There are two distinct attack goals for data poisoning: denial-of-service (DoS) poisoning and stealthy poisoning~\cite{wei2022thesis}. 

\textbf{DoS data poisoning} is a denial-of-service attack, aiming to bomb and crash the distributed learning process, making it unable to produce a usable global model. DoS poisoning assumes the adversary could compromise a large percentage of the participating clients. As a result, most of the local training data are poisoned so that the local model training will cause gradient ascent towards the adversarial objective and prevent convergence. An important countermeasure for mitigating DoS poisoning is to rely on continuous or periodic checkpoint-based server-side validation during the multiple rounds of the distributed learning process. Frequent intermediate checkpoint validation can help alert the service provider for DoS poisoning detection at the early stage. The defender can then resort to proper mitigation mechanisms, such as assigning reputation trust scores to participating clients and selecting subscribers for new distributed learning tasks based on their trust ratings. 

\textbf{Stealthy data poisoning} aims to selectively poison some learning objectives while keeping the majority of the learning task unaltered. It has two distinct features compared to DoS poisoning: causing severe prediction errors only for objects from a specific class (say a person is misclassified as a bird) while maintaining high accuracy for objects from other classes. As a result, the attack is hard to be detected. Regardless of DoS poisoning or stealthy poisoning, data poisoning can be classified into three different types based on specific poisoning strategies: label-flipping poisoning, backdoor poisoning, and adversarial example poisoning. 

\textbf{ Data poisoning by label flipping.\/} Label flipping is a low-cost yet effective data poisoning attack. It assumes that the data-at-rest and data-in-transit are secured by encryption. Data-in-use is the only vulnerable spot, i.e., when the training example is fed to the local model training at a compromised client. The attacker will replace the ground truth label of certain local training examples (source victim class) with the attack target label purposefully or randomly, while keeping the features of the training examples on the source class unchanged. Stealthy label flipping poisoning attack~\cite{tolpegin2020data} typically assumes that the percentage of comprised clients will be small, e.g., 5\% or 10\% of the total participating clients. In this case, the number of poisoned local training data examples in distributed learning is limited to avoid detection. Adaptive adversaries may design different attack strategies to increase the attack effectiveness. For example, some distributed learning services possess a stable power supply and fast WiFi connectivity. These hardware advantages enable the attackers to make themselves always available so that malicious clients can have a higher probability of being selected for joint training. Another strategy to increase the adverse effect is to launch the attack at later rounds of  distributed learning~\cite{wei2022thesis}. 

\textbf{Data poisoning by backdoor attacks.\/} 
The backdoor attack is a selective poisoning attack, which injects a backdoor tag (trigger) to a small region of some selected training examples of a specific class (source victim class), aiming to mislead the local model training to misclassify the objects of the source class by the attack target class~\cite{bagdasaryan2020backdoor}. For example, an adversary may insert some backdoor to a small region of an original training example and modifies its label to be the attack target class. By embedding the backdoor trigger through the training data, the trigger makes the local model training learn and misclassify any input example of the source class poisoned with the same backdoor trigger to the attack target class. Given that only those training examples with the backdoor trigger are poisoned and will be misclassified to the attack target class, backdoor poisoning is one type of selective data poisoning.
\cite{xie2019dba} introduces a distributed backdoor trigger with each malicious client holding a portion of the trigger. 
Backdoor attack with a universal adversarial backdoor trigger~\cite{zhao2020clean} can work with the clean label without requiring label flipping. 
Instead of injecting the backdoor trigger into a selection of training examples, one can also design and embed the poisoning trigger directly by inserting a trojan model~\cite{liu2017trojaning}. 


\textbf{Data poisoning by adversarial perturbation.\/}  Adversarial perturbation-based poisoning uses the gradient-based procedure to optimize how the training examples are poisoned to avoid detection.  Concretely, \cite{munoz2017towards} proposes an attack method that  changes the label and modifies the poisoning input based on the reverse direction of the benign training. To find the most effective scheme for adversarial data perturbation or noisy data injection, several proposals~\cite{koh2022stronger} construct an optimization 
function to search for the data that can maximize the classification error in the training set. In addition to the gradient-based optimization for adversarial perturbation-based poisoning, autoencoder~\cite{feng2019learning} is also utilized to generate adversarial perturbed poisoning of training data. 
Besides dirty label poisoning, clean-label poisoning attacks assume that the adversary cannot change the label of any training data. 
\cite{shafahi2018poison} injects imperceptible adversarial watermarks into some benign training examples with the clean label certified by a certification authority. Specifically, imperceptible adversarial watermarks from another class are injected on top of the training data such that the watermark features overtake the original features with a clean label. As a result, the poisoned training data with the adversarial watermarks can mislead the model training by minimizing the distance of the poisoned input to the instances of the attack target class. \cite{zhu2019transferable} designs targeted poisoning images with clean-label to surround the targeted image in feature space. 


\subsubsection{Model Poisoning Attacks} 
Model poisoning attempts to induce model corruption by manipulating the training procedure instead of poisoning training data. The attackers may corrupt local model updates before sending them to the federated server. \cite{fang2020local} proposes an attack that significantly deviates a global model parameter towards the inverse direction of the model update by solving the optimization problem during local model training. \cite{bhagoji2019analyzing} boosts the malicious update under only one malicious attacker by directly changing the training optimization goal and optimizes for both training loss and adversarial objective. 
Since data poisoning attacks at a compromised client eventually create a subset of local model updates sent to the model at each round of the distributed learning, model poisoning is believed to subsume data poisoning in distributed learning settings~\cite{bhagoji2019analyzing}. 
Most DoS poisoning attacks modify the training objectives using model poisoning~\cite{jagielski2018manipulating}.  
The attackers aim to prevent the convergence of the global model by applying small changes to many parameters~\cite{xie2020fall} or to cause the global model training to converge to a bad minimum~\cite{guerraoui2018hidden}.

\subsubsection{Poisoning Robustness by Detection}  Existing defense solutions against poisoning attacks rely on the assumption that the federated server in distributed learning is trusted. Hence, the primary research efforts are dedicated to detecting anomalies by separating poisoned and non-poisoned contributions. A majority of existing poisoning defense solutions are based on the detection of poisoned local model updates sent from compromised clients. 
\cite{chow2023stdlens} proposes to apply PCA on the local model updates collected over multiple rounds for each class. By leveraging the fact that only a small percentage of participating clients are compromised, there will be two distinct gradient clusters for each poisoned source class. One corresponds to benign local model updates from honest clients, and the other corresponds to the poisoned local model updates from compromised clients. 
\cite{li2021lomar} scores model updates from each remote client by measuring the relative distribution over their neighbors using a kernel density estimation method and distinguishing malicious and clean updates with a statistical threshold.
\cite{shen2016auror} proposes a method to identify the indicative features for comparison by collecting masked features from users. 
\cite{tran2018spectral} performs spectral analysis with SVD to generate two clusters for backdoor poisoning attacks. \cite{pmlr-v139-hayase21a} utilizes robust covariance estimation to amplify the spectral signature of corrupted data for detection. 
\cite{tang2021demon} proposes to decompose the input image into its identity part and variation part to perform statistical analysis on the distribution of the variation. The method 
utilizes a likelihood-ratio test to analyze the representations in each class to detect and remove the backdoor trigger. Off-line meta learning method can be trained to defend against poisoning attacks at the server to reject the poisoned model updates before global model aggregation~\cite{xu2021detecting}.


\subsubsection{Poisoning Robustness by Validation} An alternative countermeasure against poisoning attacks is server-side validation. The method assumes that the defenders have a clean validation dataset with untainted labeled data, or the clients can cross-validate each other without collusion.  \cite{hendrycks2018using} constructs a corruption matrix to separate poisoned samples using SVD. 
\cite{zhao2020shielding} requires the server to send local model updates from some clients to other clients for cross-checking.  
\cite{charikar2017learning} proposes a semi-verified learning model with a small dataset of trusted data drawn from the distribution such that the accurate extraction of information from a much larger but untrusted dataset is possible. 
\cite{cao2021fltrust} requires the service provider to collect a clean small training dataset and bootstrap the trust score for each client.
A local model update has a lower trust score if its direction deviates more from the
direction of the server model update. Then, the server normalizes the magnitudes of the local model updates such that they lie in the same hyper-sphere as the server model update in the vector space. Thus, the impact of malicious local model updates with large magnitudes can be limited.

\subsubsection{Poisoning Robustness by Certified Bound} 
The certified approach to poisoning robustness computes the centroid of each class as the certified region and removes points far away from the centroid of the corresponding class~\cite{steinhardt2017certified}. Based on randomized smoothing, \cite{rosenfeld2020certified} proves that the defender could obtain a smooth probability distribution by adding noise to the training data. \cite{levine2021deep} combines randomized smoothing with subset aggregation. However,~\cite{mehra2021robust} shows that randomized smoothing methods can be circumvented if the adversary is aware of the defense strategy. 
Under a specific notion of separability in the Reproducing kernel Hilbert space induced by the infinite-width network, 
\cite{wang2021robust} proves that training (finite-width) networks with stochastic gradient descent can be robust against data poisoning attacks. 

\subsubsection{Poisoning Robustness by Data/Model Sanitization} Unlike the previous three reactive defense approaches against poisoning attacks, the poisoning robustness by data and model sanitization promotes a proactive approach to mitigating poisoning attacks.  
For data sanitization, 
\cite{panda2022sparsefed} uses global top-k update sparsification and device-level gradient clipping to mitigate model poisoning attacks. \cite{borgnia2021strong} enforces that convex combinations of training data points are assigned convex combinations of the labels. Therefore,  the class boundaries are regularized, and small non-convex regions are removed. For model sanitization,
the dormant neurons can be pruned to weaken the poisoning impact~\cite{schuster2021you} and the adverse effect of backdoor attacks~\cite{liu2018fine}. Both methods fine-tune the model with a small clean surrogate dataset after the pruning to compensate for the utility loss. 
\cite{li2021anti} finds that the models can learn backdoored data much faster than learning with clean data. Therefore, they introduce a gradient ascent-based anti-backdoor mechanism into the standard training to help isolate low-loss backdoor examples in early training and unlearn the backdoor correlation.

\subsubsection{Analysis of Poisoning Robustness} Existing poisoning robustness techniques suffer from a number of limitations. First, the server-side detection approaches rely on the statistical robustness properties to create two distinct clusters for gradient updates from participating clients over multiple consecutive rounds of distributed training. Given that stealthy poisoning assumes that only a small percentage of compromised clients is present, most detection-based approaches assign the smaller cluster as the poisoned gradient updates and the larger cluster as the benign gradient updates. However, this assumption may not hold since cases exist in which the smaller cluster contains benign gradient updates from honest clients and the larger cluster contains poisoned gradient updates from compromised clients~\cite{wei2024gradient}. The failure to accurately distinguish benign local model update contributions from poisoned local model update contributions may result in mistakenly removing benign model updates and 
dramatically degrade the accuracy of the jointly trained global model. 
Second, additional privacy concerns need to be addressed for the validation-based approach when the federated server is not trusted or is subjected to colluding with compromised clients. Third, the certified poisoning defense bears a similar shortcoming to those in the certified evasion robustness, such as conservative bound estimation and difficulty generalizing to different learning tasks, datasets, and backbone DNN algorithms. Finally, the poisoning robustness by data and model sanitization has limited robustness~\cite{wei2022thesis} since it is challenging to determine the right amount of sanitization  sufficient to mitigate poisoning and yet small enough with minimal impact on the convergence of joint training and the accuracy of the global model. 

\subsection{Robust to Byzantine Attacks}

\label{sec3.4}

Byzantine failure in distributed systems occurs when one or more components have failed. Still, a better understanding about whether a component has failed, whether the system information is correct,  or why there is inconsistency in the system state~\cite{lamport1982byzantine} is needed. Byzantine robustness measures
the resiliency or the fault-tolerance of a distributed learning system in the presence of Byzantine failure, i.e., whether and to what extent (reliability level) the system can keep functioning even if certain components stop working. One example of Byzantine robustness is the secure aggregation of local model updates from the contributing clients in the presence of data poisoning attacks. Since distributed learning is a collaborative learning system with a geographically distributed population of participants (client devices) across different administrative
domains, it is impossible and unrealistic to guarantee that each device is trusted and reliable. 
Byzantine-robust aggregation methods should be able to sustain communications instabilities, client dropout, and erroneous model updates on top of malicious actors. 
We can classify the existing Byzantine robust aggregation techniques into two broad categories: the mechanisms for detecting and removing untrustworthy local model updates and the mechanisms for identifying honest clients.


\subsubsection{Byzantine Robustness by Removing Untrustworthy Client Data} 
One way to mitigate Byzantine failure is to detect and remove those unreliable client local training updates before performing server-side aggregation in distributed learning. It is assumed that training examples of the same class are distributed, and Byzantine attackers may only compromise some of the clients where the training data for a specific class resides. 
Hence, several solutions exploit the inconsistency between corrupted local model updates from dishonest clients and clean local model updates from honest clients. 
%
%
%
\cite{blanchard2017machine} proposes Krum, which selects one of the $m$ local models updated by the participating clients, which is most similar to other models as the global model. Krum implicitly removes those local model updates dissimilar to the chosen representative model for the given round of distributed learning. Bulyan~\cite{guerraoui2018hidden} combines Krum with a
variant of Trimmed Mean. Trimmed Mean~\cite{yin2018byzantine} aggregates the model parameters of the clients with the mean value of only those client updates close to the mean and removes the largest and smallest parameters. 
%
%
%
%
SignSGD is proposed in~\cite{bernstein2018signsgd} to aggregate sign gradients by majority vote such that no individual worker has too much power. 
%
%
%
Removing faulty client local model updates in each round may secure the per-round server-side aggregation against Byzantine failure. However, when such removal exceeds a certain threshold, it may severely burden the opportunities to learn diverse data distributions and cause survival bias in the global model. 

\subsubsection{Byzantine Robustness by Selecting Honest Clients}
Instead of detecting faulty local model updates, 
Byzantine-robust techniques can also select benign gradients and honest clients based on historical participation records.
\cite{alistarh2018byzantine} maintains a set of "good" candidates by estimating sequences and only performs aggregation using these gradient updates.
\cite{xie2019zeno} computes a score for each candidate gradient estimator using the stochastic zero-order oracle and ranks each candidate gradient estimator. The algorithm aggregates the candidates with the highest scores for synchronous aggregation. 
\cite{pan2020justinian} utilizes the historical interactions with the workers as experience 
to identify Byzantine attacks via reinforcement learning techniques.

\subsection{Robustness to Out-of-Distribution Input}

\label{sec3.2}


Out-of-Distribution (OOD) inputs are query inputs drawn from a completely different distribution than the distribution of the training data or the domain of the learning task. 
Since there is no ground-truth label for the OOD examples, the prediction output is erroneous and cannot be repaired at the test phase of the model deployment. Current countermeasure for robustness against OOD inputs is through OOD detection. By timely detecting such OOD input, the model can be retrained or fine-tuned, such as active learning or concept drift management, to handle the evolving data stream and accommodate the new class of emerging data. 
The first OOD detection method~\cite{hendrycks2016baseline} adds small perturbations to the input data by softmax-controlled input preprocessing and is often considered as the baseline approach. \cite{liang2017enhancing} improves the baseline method by combining the input noise injection with output temperature scaling by distillation temperature control. Both methods rely on properly setting the input noise amount parameter and the output temperature parameter, which are dataset-specific.
Another proposal utilizes the Mahalanobis distance with respect to the closest class conditional distribution to flag both OOD examples and adversarial examples~\cite{lee2018simple}. 
The cross-layer ensemble approach proposed in \cite{wei2020robust} can auto-verify and auto-repair adversarial examples, and auto-detect OOD inputs effectively by leveraging diverse input transformation and diverse output model prediction aggregation. 


\subsection{Robustness to Heavy-tailed and Imbalanced Data}

\label{sec3.5}

Class imbalanced data and heavy-tail data are ubiquitous in real-world machine learning applications, and the issue of class imbalance and heavy-tail data is even more severe in distributed learning since the data composition at different participating clients varies and is unknown to the aggregation server~\cite{li2019convergence}. 
Class imbalanced data refers to the problem of learning over a training dataset with a skewed distribution of training samples across different classes in supervised learning or clusters in unsupervised learning. 
Machine learning models trained over class imbalanced data may cause inferior prediction performance for the minority class compared to the prediction performance for the majority classes~\cite{van2007experimental}. 
%
%
%
%
Due to the heterogeneity of the local data distributions, there can be a significant mismatch between the local and global imbalance, i.e., the class that is a minority locally may be a majority class globally. Hence, trustworthy AI in distributed learning needs to handle two main challenges of class imbalances: global data imbalance and local data imbalance. For global data imbalance, the overall data may follow a long-tailed distribution. For local data imbalance, the number of training samples for every class on a client is highly uneven. Under the non-IID setting of distributed learning~\cite{zhao2018federated}, some clients may have no training data from a specific class. A common heuristic for dealing with long tail data is to add a compensation term to either the loss or the prediction results for the tail classes based on the training data distribution. However, such an approach is not applicable in distributed learning because only local training weights or gradients of the model can be shared by a participating client with  the aggregation server. To enable balanced sampling, loss re-weighting and gradient tuning, 
one line of work is dedicated to estimating the portion of a class using the gradient norm per class  corresponding to different classes. 
 \cite{shuaibalancefl} injects balanced sampling, feature-space augmentation smooth regularization at local client before sending the biased local model update to the server. 
\cite{shen2022agnostic} propose an agnostic constrained learning formulation, which explicitly enforces the similarity between the local empirical losses to tackle the class imbalance problem.
\cite{wang2021addressing} proposes ratio loss, but their method degrades when there is a notable mismatch between the local imbalance and the global imbalance.

\section{Trustworthy Distributed AI through Privacy-Enhancing Techniques}

\label{sec4}

Despite the default privacy in federated learning, 
recent studies 
reveal that 
simply keeping client training data local is insufficient for protecting the privacy of sensitive client training data due to gradient leakage attacks. 
Gradient leakage attacks, also known as gradient inversion attacks, are reconstruction-based inference attacks. Informally, suppose an adversary intercepts the local gradient update of a client before the server performs the federated aggregation to generate the global parameter update for the next round of distributed learning. In that case, the adversary can perform reverse engineering to steal sensitive local training data of this client by simply performing reconstruction optimization on the stolen gradient~\cite{wei2020framework}. Other types of data leakage attacks have been studied in both centralized and distributed learning environments, ranging from membership inference attacks~\cite{shokri2017membership,truex2019demystifying}, attribute inference~\cite{melis2019exploiting} to model inversion~\cite{fredrikson2015model}. In this section, We will first describe the formulation of gradient leakage attacks and then review a collection of representative privacy-enhancing techniques for increasing leakage robustness in trustworthy distributed AI.  We will conclude this section by reviewing other types of privacy leakage risks and their countermeasures.

\subsection{Gradient Leakage Attack} 
\label{sec4.1}
The gradient leakage attack discloses the proprietary client training data through two steps: (i) unauthorized read of the shared local model training parameters and (ii) unauthorized inference over the stolen local gradient from a client to reconstruct the private local training data with high confidence
%
%
To focus on privacy leakage threats during data-in-use, we assume that the training data stored at each client and the model shared between a client and its server are encrypted. 
Leakage during data-in-use may occur at either the aggregation server or any client. 

\subsubsection{Leakage at Server, at Client, and at Per Example} \label{sec:4.1.1}
At the aggregation server, an honest-but-curious adversary may collect gradient updates from each of the participating clients at each round and launch the gradient leakage attack, which will infer the sensitive local training data used to produce the local model update of a client prior to performing federated stochastic gradient descent (SGD) at each round. In the meantime, they reliably coordinate the federated learning process without damaging the workflow or changing prediction results. We refer to this server-side per-client gradient leakage attack as \textbf{leakage at server}. 
The gradient leakage attack can also happen at individual clients if a semi-honest agent on the client gains unauthorized access to the local model update upon the completion of the local training and before the client encrypts it and sends it to the server. We refer to this client-side per-client gradient leakage attack as \textbf{leakage at client}. The main difference between leakage at server and leakage at client lies in (i) the attack location and (ii) the attack coverage in a given round. Leakage at server can be performed on any participating clients in a given round. In contrast, leakage at client can only reconstruct the local training data of the given client. 
The third type of gradient leakage attack is called  \textbf{leakage at per example}. Unlike leakage at server and leakage at client,  leakage at per example is performed on the per-example gradient before performing the local SGD at each local iteration during the local model training at a client. 

\subsubsection{Attack Formulation and Algorithm}
The gradient leakage attack can be formulated as a reconstruction learning function $A: Z_{x} \rightarrow x_{rec}$, with the following attack objective, 
where $Z_{x}$ denotes the leaked gradient corresponding to the training data $x$, and $x_{rec}$ is the reconstructed training data corresponding to $x$:
\begin{align}
   \min ||\nabla_{x_{rec}} f-Z_{x}||_2 \quad 
    s.t. \quad & ||x - x_{rec}||_2 \approx 0.
\end{align}
The optimization goal of the gradient leakage attack is to iteratively modify $x_{rec}$ by minimizing the distance between the gradient of the reconstructed input $\nabla_{x_{rec}} f$ and the leaked gradient value $Z_{x}$ such that the reconstructed input $x_{rec}$ gradually becomes visually close to the private training data $x$ and eventually exposes the training example $x$ with high confidence as they become almost identical: $x_{rec} \approx x$. 
Concretely, 
the attack algorithm $A: Z_{x} \rightarrow x_{rec}$ starts with a dummy seed $x^{0}_{rec}$, 
generates the attack gradient $\nabla_{x_{rec}}f$, and backpropagates using the local model. It modifies the seed input by minimizing the loss $D^{\tau}$, defined by the Euclidean distance between the gradient $\nabla_{x_{rec}}f$ for the reconstructed data and the leaked gradient $Z_{x}$ at attack step $\tau$. 
The attack ends when the loss is smaller than a pre-defined threshold or reaches the pre-defined $\mathbb{T}$ termination steps.
For leakage at server and leakage at client, the leaked gradient of client $i$ is the accumulated result after the local training over the local training set $X$ at round $t$, denoted by $Z_{X}$. The attack is to reconstruct training data $x_{rec}$ that is visually close to one of the private training examples in $X$. 
\begin{align}
   \min ||\nabla_{x_{rec}} f-Z_{X}||_2 \quad 
    s.t. \;  \exists_{x \in X} ||x - x_{rec}||_2 \approx 0.  
\end{align}
%
%
%
%

\subsubsection{Attack Optimizations}
Gradient leakage attacks are investigated from multiple perspectives in the literature, aiming to optimize the attack efficiency or to expand to different data modalities, learning tasks, and architectures.

\textbf{Recursive attack.} The early attempt of gradient leakage~\cite{aono2017privacy} brings theoretical insights by showing provable reconstruction feasibility on a single-layer DNN model. The attack infers private data based on the proportionality between the training data and the gradient updates. Given $wx+b=z$ of one layer, $\frac{\nabla l}{\nabla w} = \frac{\nabla l}{\nabla z}x^{transpose}, \frac{\nabla l}{\nabla b} = \frac{\nabla l}{\nabla z}$ such that $x^{transpose}=\frac{\nabla l}{\nabla w}/\frac{\nabla l}{\nabla b}$. \cite{zhu2020r} combines the forward and backward propagation and formulates the problem as solving a system of linear equations.

\textbf{Initialization seed.} Given a gradient attack algorithm, the probability and speed of the attack convergence depend heavily on the bootstrapping initialization seed. The attack in~\cite{zhu2019deep} is unstable. \cite{wei2020framework} proves that geometric initialization improves both the convergence probability and speed compared to random seed-based bootstrapping~\cite{zhu2019deep}. In comparison, the gradient leakage attack in~\cite{geiping2020inverting} requires a much longer time to perform successful reconstruction, usually $2\sim3$ times longer than the default attack termination condition used in DLG~\cite{zhu2019deep}, regardless of whether to use the initial random seed as in DLG or the patterned seed as in the CPL attack~\cite{wei2020framework}.
%
%
%
%
To bypass the gradient obfuscation by noise addition and gradient compression, \cite{yue2022gradient} encodes the initialized input into low-dimensional representation and decodes it back to the image domain after the attack optimization.

\textbf{Attack loss function optimization.}  
   As the attack optimization function, L-BFGS is used in early attack algorithms~\cite{zhu2019deep}. Given that L-BFGS requires third-order derivatives,  this leads to challenging optimization problems for network activation functions, such as ReLU, to converge under the second-order derivatives. As a result, \cite{geiping2020inverting} replaces L-BFGS with Adam in the attack algorithm. However, this could significantly increase the attack cost. 

  \textbf{Training data resolution \& data modality.} While the early algorithm for gradient leakage attack outlined in~\cite{zhu2019deep} states that the reconstruction cannot succeed when the resolution of the training image data is larger than 64*64, several follow-up efforts~\cite{yin2021see} show the feasibility of succeeding gradient leakage attacks on DNN trained over 
     large-scale and high-resolution datasets like ImageNet and X-Ray. Using more efficient attack algorithms through different optimizations, gradient leakage attacks can succeed with a much smaller number of reconstruction iterations.~\cite{dimitrov2022lamp} extends the attack to text data. 

\textbf{Batch size, learning rate, and client disaggregation.} 
The early gradient leakage attack algorithm in
\cite{zhu2019deep} uses separate weights and sub-models for each training example (batch size of one) to show the reconstruction inference by reverse engineering. The same logic applies to batch sizes of up to 8. The attack algorithm with initial seed optimization in \cite{wei2020framework} shows that the reconstruction of multiple images from the averaged gradient is possible (for a batch size of 16). The loss-function optimized attack algorithm in \cite{geiping2020inverting} shows the feasibility of using the same logic for the gradient-based reconstruction of an arbitrarily large batch of training data, e.g., batch size of 100.
By utilizing regularization terms, \cite{yin2021see,huang2021evaluating} can recover the original private training data through reconstruction inference with a batch size of over 30. 
The learning rate in the attack algorithm may also impact the attack efficiency. \cite{wei2020framework} shows that gradient leakage attacks could fade away after some local training. A possible explanation is that the attack in~\cite{zhu2019deep} uses a fixed attack learning rate. Then, as the gradient value becomes smaller as the learning round progresses and approaches the termination round, the attack becomes less and less effective.
Furthermore, \cite{lam2021gradient} disaggregates the averaged gradient and reconstructs the binary user participant matrix. They formulate the problem as matrix factorization and solve it using standard mixed-integer programming. 

       
        

\textbf{Vertical federated learning.} Recent studies show that gradient leakage attacks can cause equally severe privacy violations in vertical federated learning systems. 
Concretely, 
        \cite{jin2021cafe} leverages the novel use of data index and internal representation alignments in vertical federated learning to optimize gradient leakage attacks.  \cite{fu2022label} focus on label inference in vertical federated learning by exploiting the bottom model trained locally by each participant.

        

  \begin{table}[t]
\centering
\caption{\small Comparison of the representative gradient leakage attack methods under leakage at per example.}
 \vspace{-0.3cm}
\scalebox{0.85}{
\small{
\begin{tabular}{|cc|c|c|c|}
\hline
\multicolumn{2}{|c|}{}                                    & MNIST &  CIFAR10 & LFW   \\ \hline
\multicolumn{1}{|c|}{\multirow{3}{*}{DLG~\cite{zhu2019deep}}}     & attack success rate      & 0.686         & 0.754   & 0.857 \\ \cline{2-5} 
\multicolumn{1}{|c|}{}                         & reconstruction quality (SSIM)  & 0.985         & 0.998   & 0.997 \\ \cline{2-5} 
\multicolumn{1}{|c|}{}                         & reconstruction attack iteration & 18.4          & 114.5   & 69.2  \\ \hline
\multicolumn{1}{|c|}{\multirow{3}{*}{GradInv~\cite{geiping2020inverting}}} & attack success rate      & 1               & 0.985   & 0.994 \\ \cline{2-5} 
\multicolumn{1}{|c|}{}                         & reconstruction quality (SSIM)  & 0.989         & 0.998   & 0.998 \\ \cline{2-5} 
\multicolumn{1}{|c|}{}                         & reconstruction attack iteration & 846          & 2135    & 1826  \\ \hline
\multicolumn{1}{|c|}{\multirow{3}{*}{CPL~\cite{wei2020framework}}}     & attack success rate       & 1                 & 0.973   & 1     \\ \cline{2-5} 
\multicolumn{1}{|c|}{}                         & reconstruction quality (SSIM)  & 0.99         & 0.998   & 0.998 \\ \cline{2-5} 
\multicolumn{1}{|c|}{}                         & reconstruction attack iteration & 11.5        & 28.3    & 25    \\ \hline
\end{tabular}
}}
 \centerline{\includegraphics[scale=.55]{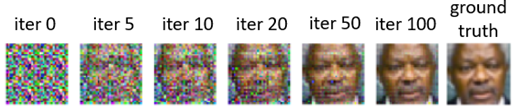}}
\label{table:attack_compare}
 \vspace{-0.5cm}
\end{table}

\subsubsection{Analysis of Gradient Leakage Attacks}
We perform a set of experiments to compare three representative attack algorithms (DLG~\cite{zhu2019deep}, GradInv~\cite{geiping2020inverting}, and CPL~\cite{wei2020framework}) for leakage at per example on 100 examples of MNIST, CIFAR10, and LFW,  respectively.  \textbf{Table~\ref{table:attack_compare}} shows the results. 
The attack effectiveness is measured by the attack success rate (ASR), the reconstruction quality in SSIM, and the number of reconstruction iterations required for the attack to succeed. 
In comparison, CPL is the most efficient attack algorithm with the shortest reconstruction time and highest reconstruction quality and attack success rate. DLG takes 1.5$\times$ to 4.5$\times$ longer in the total number of reconstruction iterations to succeed compared to CPL. GradInv takes much longer to succeed for all three datasets than CPL and DLG. DLG has the lowest attack success rate in comparison. For MNIST, GradInv and CPL have 100\% ASR, whereas DLG has an ASR of 0.686. For CIFAR-10, GradInv has an ASR of 0.985, CPL has an ASR of 0.973, whereas DLG has an ASR of 0.754. For LFW, CPL has the highest ASR of 100\%, GradInv has an ASR of 0.994, and DLG has an ASR of 0.857. The visualization at the bottom of \textbf{Table~\ref{table:attack_compare}} shows the iterative attack process starting from the patterned initialization seed 
and demonstrates the efficiency of the CPL attack on an LFW instance.

\subsection{Robustness by Privacy-Enhancing Techniques} 

 \label{sec4.2}


\subsubsection{Leakage Robustness by Differential Privacy (DP)}  DP is a de facto standard for publishing deep learning models with a statistical privacy guarantee. \cite{dwork2014algorithmic} states that a randomized mechanism $\mathcal{M}$: $\mathcal{D}\rightarrow \mathcal{R}$ satisfies ($\epsilon,\delta$)-DP if for any two input sets $A \subseteq \mathcal{D}$ and $A'\subseteq \mathcal{D}$, differing with one entry: $||A-A'||_{0}=1$, Equation~\ref{equa:dp_centrailized} holds with $0 \leq \delta < 1$ and $\epsilon>0$.
\begin{equation}
\Pr(\mathcal{M}(A) \in \mathcal{R}) \le e^{\epsilon}\Pr(\mathcal{M}(A') \in \mathcal{R}) + \delta,
\label{equa:dp_centrailized}
\end{equation}
where $\mathcal{D}$ is the domain of input data and $\mathcal{R}$ is the range of all possible output.
In traditional deep learning with DP, the gradient update is clipped by a pre-defined clipping bound $C$, where the clipping parameter caps the $l_2$ norm of the gradient to estimate the sensitivity $S$. The sensitivity indicates the maximum amount that the function varies when a single input entry is changed. 
Then Gaussian noise calibrated to the clipping bound $C$ and a pre-defined noise scale $\sigma$ is added to sanitize the gradient and perform SGD at each iteration~\cite{abadi2016deep}. The per-example noise injection ensures instance-level DP, meaning that one cannot tell from the trained model if a specific data piece is used in the training.

Most existing approaches to federated learning with DP~\cite{mcmahan2017learning,geyer2017differentially} 
apply clipping and noise injection on per-client per-round local model update. The noise injection is done on the client update for server aggregation, regardless of whether the DP noise is injected at the federated server prior to performing global aggregation~\cite{mcmahan2017learning} or at each of the participating clients prior to encrypting the gradients and sending them to the federated server~\cite{geyer2017differentially}.
The local training is the same as the non-private training, and the sanitization process only occurs after the completion of local training. 
The per-client per-round noise makes server-side aggregation differentially private and the resulting global model differential private only at the client level. However, client's local training is not protected by DP noise. Therefore, gradient leakage attacks can happen anytime before the sanitization. 
By comparison, \cite{wei2021gradient} considers per-example per-iteration DP noise injection during the client's local training. The method by design closely follows the conventional DP definition at the instance level~\cite{dwork2014algorithmic} and ensures that local SGD is differentially private.
Due to composition theorem and post-processing, the globally trained model is differentially
private at both the instance level and client level.

While conventional DP noise injection concerns fixed noise with fixed DP parameters, we find that not all versions of DP noise are gradient leakage resilient~\cite{wei2023model}. This 
raises the question of how to choose the parameter setting from these many versions under a given privacy budget such that the resulting model could prevent gradient leakage and yet does not suffer much on the accuracy performance. 
\cite{wei2023securing} offers one possible solution by injecting dynamic per-example DP noise. With noise calibrated with the decaying noise scale and the actual $l_2$ norm of the gradient update, the proposed approach is able to offer gradient leakage resilience with high utility guarantee.

Besides traditional DP, federated learning with local differential privacy (LDP) extends the LDP concept and techniques~\cite{kasiviswanathan2011can},   
like RAPPOR~\cite{erlingsson2014rappor} and Condensed LDP~\cite{gursoy2019secure} 
to protect user identity. 
Although the method~\cite{truex2020ldp}  
perturbs the per-client gradient updates or locally splits and shuffles the weights~\cite{sun2021ldpfl} before sharing with the federated server, it does not generate a global model with DP guarantee as the $\epsilon$ in LDP is defined over the local data record rather than the global data space. Also, the per-client per-round noise injection makes it vulnerable to leakage at per example. 
 While the traditional DP mechanism is based on the Gaussian mechanism or randomized response, other noise injection approaches like Binomial mechanism~\cite{agarwal2018cpsgd} and relaxations of the DP definition are explored in federated learning: 
 Gaussian DP~\cite{zheng2021federated} and teacher-student model~\cite{papernot2018scalable}. 

\subsubsection{Leakage Robustness by Secure Multiparty Computation  (SMPC)}  SMPC~\cite{canetti2000security}) is a subfield of cryptography that allows creating methods to jointly compute a function using inputs from different parties without
revealing those inputs neither to each other nor to the central
server~\cite{bonawitz2017practical}. 
 \cite{mohassel2017secureml} allows data owners to train various models on their joint data without revealing any information beyond the outcome by 
 encrypting and/or secret-sharing their data among two non-colluding servers.
\cite{kairouz2015secure} proposed to combine SMPC with DP. It allows each party to randomize its bit and share the privatized version with the other parties to prevent inference from a central observer interested in computing a separate function on all the clients' bits. However, SMPC mainly focuses on secure sharing between a client and the federated server. It can protect against {\bf leakage at sever} because SMPC secures the per-client local training update received from a client. However, SMPC may not secure {\bf leakage at client} and {\bf leakage at per example} (recall Section~\ref{sec:4.1.1}). 
First, leakage at client can still be launched successfully at a client before the local model updates (gradients) are cryptographically processed by secret share generation. 
Second, leakage at per example cannot be protected by SMPC because the local SGD is performed on raw gradients of all examples in each minibatch per local iteration. 
%
%
%

\subsubsection{Leakage Robustness by Homomorphic Encryption (HE)}
HE allows arithmetic operations to be directly performed on ciphertexts and thus provides strong privacy protection as it allows training a model on an encrypted dataset.
HE techniques can be categorized into: 1) fully HE~\cite{gentry2009fully} and 2) partially HE~\cite{rivest1978method, paillier1999public,elgamal1985public}. Partially HE allows only a single operation to be performed on cipher text, and fully HE can support multiple operations. 
Despite the recent advances in HE, performing arithmetic on encrypted numbers comes at the cost of memory and processing time. 
Furthermore, it cannot withstand collusion between the server and multiple participants. 
\cite{zhang2020batchcrypt} tries to reduce the encryption and communication overhead by encoding a batch of quantized gradients into a long integer instead of individual gradients with full precision.

\subsubsection{Leakage Robustness by Trusted Execution Environment  (TEE)}
TEE~\cite{subramanyan2017formal} is defined as a trusted platform for executing attested and verified code by securing enclaves. TEE enables the user to remotely verify the confidentiality and integrity of the data and code used in the enclave running on an untrusted host. However, distributed AI systems must be designed and implemented to accommodate different implementations of TEE, e.g., Intel SGX, AMD SEV, and ARM TrustZone, which requires non-trivial, error-prone program transformations. 
For example, TEE in~\cite{subramanyan2017formal}  protects the data 
in the enclave 
against injecting false training results by enabling an isolated, cryptographic electronic structure and permitting end-to-end security. For leakage at server attacks, supporting TEE runtime at the federated server can be effective~\cite{mo2021ppfl}. However, to rely on TEE to mitigate inference attacks at client, such as leakage at client and leakage at per example, supporting TEE runtime at each client is mandatory, which may not be feasible for thin edge clients with limited computation resources.

\subsubsection{Leakage Robustness by Selective Data Sharing with Random Noise} \cite{shokri2015privacy} is the first proposal for privacy-preserving federated learning, which uses selective and random sharing of model parameters with randomized noise for privacy protection. Later studies have shown that gradient pruning 
can be partially leveraged to remove the essential information needed for reconstruction~\cite{zhu2019deep,wei2020framework}. 
Several recent efforts have been dedicated to selectively sanitizing the raw gradients to reduce the amount of random noise injected. \cite{sun2021soteria} shows that applying gradient compression only on one model layer may be sufficient to mitigate gradient leakage for certain scenarios. 

\subsubsection{Leakage Robustness by Leveraging Synthetic Data}
Synthetic Data generation technologies are widely recognized as a potential methodology that can be used to create and share realistic data freely. Developing safe and utility-preserving synthetic data generation techniques can alleviate both privacy concerns with sharing raw data and ethical AI issues related to bias~\cite{gursoy2018utility,lu2023machine}.  
For example,  \cite{wan2022defense} generates synthetic samples that are distinct from the original ones, so the reconstruction cannot obtain the actual training data. 

\subsubsection{Analysis of Privacy-Enhancing Methods}

We here provide an analytical review of a suite of privacy-enhancing techniques and their robustness in mitigating against gradient leakage attacks, ranging from leakage at server and client to per example (recall Section~\ref{sec:4.1.1}). First, federated learning with server-side DP guarantee only ensures that the global SGD is differentially private~\cite{mcmahan2017learning,geyer2017differentially}. It does not ensure that the local SGD performed at each participating client is differentially private. As a result, it lacks robustness against leakage at client and leakage at per example~\cite{wei2021gradient}.
Second, 
SMPC is a cryptographic technique for enhancing privacy in multiparty communication and computation systems, such as securing per-client local model updates sharing with a remote and possibly untrusted aggregation server in distributed learning systems. Hence, SMPC offers strong robustness against leakage at server and other types of inference attacks while having minimal impact on the accuracy of the global model compared to DP-based solutions~\cite{truex2021tsc}. However, the main bottleneck of SMPC is the high communication cost.
Also, SMPC is not designed to protect against inference attacks to privacy at local clients, such as leakage at client and leakage at per example. 
Third, HE and TEE are cryptographically capable of preventing inference attacks at client and server, as long as the server and clients are capable of supporting TEE or HE, respectively. For instance, each client must install TEE and ensure that both the local model training and local training data are hosted in the TEE enclave. An alternative solution is to have the local training data encrypted at rest, and the enclave has direct access to the data with TEE~\cite{ohrimenko2016oblivious}  to prevent the exploitation of any side channels induced by disk, network, and memory access patterns using SMPC.
In these scenarios, TEE is robust against leakage at client and leakage at per example. Similarly, if HE is supported at both server and every client to enable every local SGD to be performed using HE, then HE is robust against leakage at client and leakage at per example. However, the cost of running TEE and HE at edge clients with limited resources can be challenging. Finally, selective sharing with random noise suffers from the manual tuning of the Gaussian noise threshold, which is dataset and training algorithm dependent~\cite{wei2021gradient_tifs}. 
Another main challenge facing these privacy-enhancing schemes is the requirement for simultaneous coordination and compliance of all participants during the entire training process (See compliance issues in Section~\ref{sec6}). The observatory will embrace inclusive and equitable data access by leveraging different privacy-enhancing techniques and exploring different ways to combine them with other emerging approaches to safeguarding the privacy and security of data and access. 
\textbf{Table~\ref{table:robustness_techniques}} summarizes the privacy-enhancing techniques in the presence of gradient leakage at server, leakage at client, and leakage at per example, as well as training data leakage at model prediction phase (leakage at inference). 

 \begin{table}[t]
\centering
\caption{\small Overview of robustness for privacy-enhancing techniques under different types and locations of privacy leakage. }
\vspace{-0.3cm}
\scalebox{0.90}{
\small{
\begin{tabular}{|c|c|c|c|c|}
\hline
                          & \begin{tabular}[c]{@{}c@{}} robustness to \\ leakage  at per example \end{tabular}  & \begin{tabular}[c]{@{}c@{}}robustness to \\ leakage  at client \end{tabular}   &   \begin{tabular}[c]{@{}c@{}}robustness to \\ leakage  at server\end{tabular}  & \begin{tabular}[c]{@{}c@{}} robustness to \\ leakage  at inference \end{tabular} \\ \hline
DP at aggregation server  & no                     & no                 & yes               & yes                  \\ \hline
DP at client after local training      & no                     & yes                & yes               & yes                  \\ \hline
DP at per example         & yes                    & yes                & yes               & yes                  \\ \hline
LDP                       & no                     & yes                & yes               & no                   \\ \hline
SMPC                      & no                     & no                 & yes               & no                   \\ \hline
HE at global aggregation  & no                     & no                 & yes               & no                   \\ \hline
HE at local training \& global aggregation      & yes                    & yes                & yes               & no                   \\ \hline
TEE at global aggregation & no                     & no                 & yes               & no                   \\ \hline
TEE at local training \& global aggregation    & yes                    & yes                & yes               & no                   \\ \hline
\end{tabular}
}}
\label{table:robustness_techniques}
\vspace{-0.4cm}
\end{table}

\subsection{Robustness to other Privacy Violations}

\label{sec4.3}

\subsubsection{Membership Inference Attack} Membership inference attack aims to infer whether a test data sample is a member of the training set based on the prediction result produced by a pre-trained model during model deployment~\cite{shokri2017membership}. 
Membership inference attacks can happen to models trained in a centralized setting or federated learning where training data is geographically distributed across a population of clients~\cite{truex2019demystifying}.  
\cite{melis2019exploiting} introduces the first gradient-based membership inference attack in federated learning. 
The authors show that the non-zero gradients of the embedding layer of a recurrent neural network model trained on text data can reveal which words are in the training batches of the honest participants. The membership vulnerability is because the embedding is updated only with the words that appear in the batch, and the gradients of the other words are zeros. 
\cite{nasr2018comprehensive} tempers with the federated training process and intentionally updates the local model parameters to increase the loss on the target data record.
If the target data record is a member of the training set, applying gradient ascent on the record will trigger the model to minimize the loss of this record by gradient descent, whose sharpness and magnitude is much higher than performing gradient ascent on data records that are not members of the training set. 
Different proposals have been put forward for enhancing robustness against membership inference, including prediction confidence masking~\cite{jia2019memguard}, 
droupout~\cite{leino2020stolen}, 
and model compression~\cite{wang2021against}. 
However, these techniques can provide only limited robustness against the membership inference, 
and none can eliminate the privacy threat completely or at a high defense success rate~\cite{truex2019effects}.

\subsubsection{Feature Inference Attack} 
Instead of leaking the membership of test data based on the prediction results, the feature or property inference attack~\cite{melis2019exploiting} attempts to extract some underlying properties or statistics of the training dataset for a specific client, which are uncorrelated to the training task. 
Besides targeted at horizontal federated learning, \cite{luo2021feature} further extends the feature inference attack to vertical federated learning and \cite{pasquini2021unleashing} to split learning. 
The discussion on defense against attribute inference and model inversion is rather limited. \cite{jia2018attriguard} adds adversarial noise to attribute value to alter the probability distribution in attribute inference. 
\cite{li2022ressfl} derives a resistant feature extractor via attacker-aware training to initialize the client-side model to defend against model inversion attacks.

\subsubsection{Model Extraction Attack} Unlike the membership or attribute inference attack through data reconstruction, the 
model extraction attack creates a substitute model that behaves similarly to the model under attack. The substitute models are supposed to match the accuracy of the model both within the learning task~\cite{tramer2016stealing} and out of the learning task, e.g., making the same mistake given the same input~\cite{jagielski2020high}. The adaptive adversary will avoid anomaly detection during the model extraction by optimizing the attack to use as few queries as possible.
%
%
%
%
Apart from creating substitute models to learn the behavior of the target model, \cite{wang2018stealing} extracts hyper-parameters in the objective function. 
For model extraction defense, \cite{kariyappa2020defending} injects incorrect predictions for out-of-distribution queries to degrade the accuracy of the attacker's clone model. In a similar spirit, \cite{orekondy2020prediction} actively perturbs predictions targeted at poisoning the training objective of the attacker. 
PRADA~\cite{juuti2019prada} analyzes the distribution of successive queries from a client and identifies deviations from the Gaussian distribution. \cite{jia2021entangled} classifies data sampled from the task distribution and data encoding watermarks. An adversary attempting to remove watermarks entangled with legitimate data is also forced to sacrifice performance on legitimate data. 

\subsection{Interplay between Privacy-Enhancing Techniques and Poisoning Robustness}

\label{sec:4.4}

\begin{figure}[t]
\begin{minipage}[t]{0.49\linewidth}
  \centerline{\includegraphics[scale=.61]{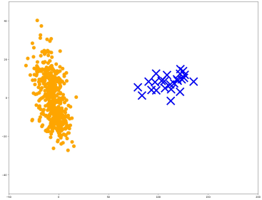}}
 \vspace{-0.1cm}
\subcaption{\small non-private.}
 \label{fig:poisoning_nonprivate}
 \end{minipage}
\begin{minipage}[t]{0.49\linewidth}
 \centerline{\includegraphics[scale=.445]{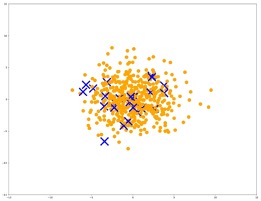}}
 \vspace{-0.1cm}
\subcaption{\small under differential privacy.}
 \label{fig:poisoning_dp}
  \end{minipage}
  \vspace{-0.4cm}
 \caption{\small PCA at the server on the local model update shared from the client on CIFAR10. The victim class is 1 (Car), and the target class is 9 (Truck). Yellow dots are benign gradient updates, and blue crosses are poisoned ones~\cite{wei2024demystifying}. }
  \label{fig:poisoning}
  \vspace{-0.4cm}
  \end{figure}

Robustness-enhancing techniques against data poisoning (recall Section~\ref{sec3}) are geared at finding ways to distinguish the benign gradients obtained from the honest clients from the poisoning gradients from the compromised clients. Interestingly, some 
privacy-enhancing techniques against privacy leakage attacks, represented by DP, leverage data sanitization methods to perturb the raw gradients from honest clients for gradient leakage attack prevention. In this subsection, we discuss the interplay and the potential synergy between privacy-enhancing techniques and poisoning mitigation techniques. First, the adverse effect of data poisoning can be reduced in a differentially private federated learning system for two reasons: (1) for a given class, the per-client gradients produced from the poisoned training examples tend to have certain statistical properties that are distinct from the benign gradients obtained from the non-poisoned training data. However, the gradients sanitized by DP noise tend to reduce the adverse effect of poisoning gradients by decreasing the statistical difference between poisoned gradients and benign gradients~\cite{wei2024demystifying}.  
\textbf{Figure~\ref{fig:poisoning_nonprivate}} shows an example of poisoned gradients v.s. non-poisoned gradients received by the aggregation server in a non-DP federated learning system. We observe two distinct clusters representing poisoning and benign gradients received by the aggregation server from compromised clients and honest clients. \textbf{Figure~\ref{fig:poisoning_dp}} shows the results of running a differentially private federated learning system for the same learning task and dataset under the same poisoning attack. We observe that under DP noise sanitization, both poisoned and benign gradients are perturbed by DP-controlled noise injection. As a result, the adverse effect of poisoned gradients after DP sanitization is no longer distinct from those perturbed benign gradients. However, poisoning attacks can be tailored to circumvent the DP-controlled noise~\cite{ma2019data}.
DP has also been used for evasion robustness~\cite{lecuyer2019certified} and backdoor robustness~\cite{nguyen2022flame}.
In addition, \cite{liu2021privacy} integrated HE with zero-knowledge proof and secure aggregation to defend against poisoning attacks by ensuring gradient indistinguishability. 
\cite{ma2022shieldfl} proposed to mitigate model poisoning by combining HE with the secure cosine similarity method to measure the distance between two encrypted gradients for Byzantine-robust aggregation. However, \cite{mahloujifar2019data} demonstrates that the multiparty learning process might suffer from poisoning attacks, and there has yet to be a countermeasure to date. 

\section{Trustworthy Distributed AI in Distributed bandits}

\label{sec5}

Distributed bandits are also susceptible to both privacy leakage and adversarial attacks. In this section, we review the design of privacy-enhancing methods and robustness-enhancing algorithms in the context of privacy violation, heavy-tailed data, or adversarial data corruption for distributed bandits.

\textbf{Robust distributed bandits against privacy intrusion.} 
In federated and multi-agent bandits, a differentially private system must ensure that the policy concerning the rewards obtained from the previous rounds is differentially private when the system interacts with a new user at the new round~\cite{tossou2016algorithms}. Specifically, messages concerning the reward sequence are differentially private, such that any policy using these messages is differentially private due to post-processing~\cite{dwork2014algorithmic}.
Distributed multi-agent multi-armed stochastic bandit is made differentially privacy~\cite{dubey2020differentially} by extending the method in~\cite{wang2019distributed} with a tree-based DP approach from~\cite{dwork2010differential}.
Unlike distributed private deep learning that explores Gaussian mechanism~\cite{dwork2014algorithmic}, distributed bandits usually achieve DP by Laplace mechanism~\cite{tossou2017achieving}. 

\textbf{Robust distributed bandits against heavy-tailed rewards.} 
The heavy-tail problem also exists in the distributed bandit setting and is first brought out by~\cite{bubeck2013bandits} and solved with robust mean estimators called Robust-UCB to obtain logarithmic regret. 
Near-optimal algorithm under heavy tails is also explored~\cite{shao2018almost} with a mean of medians-based statistical estimator. 
\cite{dubey2020cooperative} studies the heavy-tailed stochastic bandit problem in the cooperative multi-agent setting and introduces a decentralized multi-agent algorithm for the cooperative stochastic bandits that incorporates robust estimation with a message-passing protocol. 
However, most of these algorithms for heavy-tailed decision-making in the bandit problem (1) focus on the centralized setting and (2) rely on the robust mean estimators to construct tight confidence intervals for reward estimation, which typically require excessive communication in the multi-agent setting.

\textbf{Robust distributed bandits against Byzantine and adversarial corruptions.} 
Robust estimation with adversarial corruption has a rich history in bandit literature. 
In the distributed setting, cooperative estimation has been explored for sub-Gaussian stochastic bandits using a running consensus protocol~\cite{landgren2016distributed_CDC}. 
Contrasted to cooperative settings, multiple agents can compete for arms~\cite{bistritz2018distributed}. 
\cite{branzei2021multiplayer} discusses cooperative and competitive agents regarding when the players compete in a zero-sum game and when their interests are aligned. 
While the word ``adversarial" in most bandit literature concerns distribution change, evasion attacks~\cite{jun2018adversarial} 
and poisoning attacks~\cite{liu2019data} are brought to the stochastic bandits by lowering the reward
of arms. 
Besides adversarial corruption, the distributed bandits are also vulnerable to Byzantine attacks. \cite{dubey2020private}  offers  upper confidence bound in a cooperative setting when the agent can provide stochastically incorrect information. 
Model misspecification is another type of threat in the bandit problem~\cite{foster2020adapting}.

\section{Trustworthy Distributed AI through Fairness and Governance}

\label{sec6}

We have discussed trustworthy distributed AI through AI robustness and AI privacy. 
Fairness is another crucial dimension of trustworthy AI. It is a key component for establishing responsible AI and ethical AI guidelines and frameworks to ensure that AI algorithms and their deployment for real-time decision-making are secure, privacy-preserving, safe,  and fair. 
In the distributed AI decision-making process, 
 different types and sources of biases can cause unfairness. On the one hand, biases in data may lead AI algorithms to make unfair predictions or recommendations. On the other hand, AI algorithms may further aggravate the unfairness brought into the AI algorithms through biases in data. 
We dedicate this section to discussing trustworthy AI through fairness first and then discuss the potential of leveraging data governance and model governance frameworks to empower stronger synergy among AI robustness, AI privacy, and AI fairness.

\subsection{Trustworthy Distributed AI through Fairness}

In this section, we focus on two potential sources of unfairness in distributed learning systems: (1) biases in data, which can skew the learning of AI algorithms, and (2) biases in algorithms, which may be optimized with specific nuances/subtleties that are sensitive to AI fairness measures and hence handicap/hinder them from making fair decisions. 

\subsubsection{Biases in Data}
Biases in data could come from at least two sources: data preprocessing or data collection.  

\textbf{Preprocessing bias.} The preprocessing of the data fed to the training of an AI model could incur bias. Simpson's paradox~\cite{blyth1972simpson} is a general type of preprocessing bias caused by a statistical phenomenon: when a population dataset is partitioned into subpopulations, an association between two variables in a population may emerge, disappear, or reverse.
\cite{kievit2013simpson} shows that Simpson's paradox may occur when inferences are drawn across different levels of explanation, such as from population to subgroups or from subpopulations to individuals. Such phenomenon in the geospatial analysis is called modifiable areal unit problem (MAUP)~\cite{gehlke1934certain} and in time series partition called longitude data fallacy~\cite{barbosa2016averaging}. 
The MAUP problem shows a statistical biasing effect may be present when data samples in a given spatial area represent information such as density in a given area~\cite{xu2014sensitivity}. 

One way to mitigate data biases in preprocessing is to scrutinize, regroup and resample the data as much as we can. Therefore, we can identify and use statistical markers to explain the paradox. Explicit modeling of situations in which the paradox might occur can prevent biased/incorrect interpretations of the data. The modeling can also enhance our understanding of the multiple different conclusions that can be drawn from the different categorizations of data. In the meantime, the modeling can help us to choose the best data viewpoint that gives a fair representation of the world. 

\textbf{Data collection bias.} The notion of bias is context-driven. Example context can be defined by demographics, sex, age, race, income, education, and employment. The context-related data skewness often exists during any transaction-based data collection process. For example, purchase transactions for electronics may be skewed by sex since more men buy electronics than women. The nursing school admission may show some skewness by age and sex since the percentage of young women being admitted is high. 
Such data skewness indicates some sort of social and population bias in data, which may affect how the data is analyzed and measured~\cite{tufekci2014big}. 

One practical approach to mitigate data collection biases is associating each model with an explainable interpretation, including explaining the context-based prediction by providing statistical markers on the context-relevant data skewness. Explicit modeling of situations in which biases in data may be present will not only avoid misconceptions of the data but also results in a deeper understanding and a better explanation of the fairest interpretation of what data tell us.

\subsubsection{Biases in Algorithms} 

In addition to biases in data, the prediction models trained by learning algorithms can be biased due to certain algorithm design choices, such as optimization strategies, regularization functions, or biased estimators~\cite{baeza2018bias}. 
Such biases in algorithms may further amplify the disparity between the majority and minority classes in a skewed data distribution. For example, the attack-resilient mechanisms and privacy-enhancing techniques could aggravate the bias. Recent research in~\cite{bagdasaryan2019differential}  
has shown federated learning with DP may aggravate the data skewness induced disparity with respect to accuracy for minority subpopulations.

To circumvent biases in algorithms, explainable AI techniques should create context-based and fairness-aware interpretations of the algorithmic decisions. One example is to explain data skewness by different population categorizations using different fairness-sensitive context variables. 
The fairness-sensitive context variables can be provided through data and model governance policies and specifications (see Section~\ref{sec6.2}). 

\subsubsection{Fairness Definition}
The different types of biases call for a unified definition of fairness. However, the fact that no universal definition of fairness exists shows the difficulty of solving this problem \cite{hutchinson201950}.  
For example, \cite{dwork2012fairness,kusner2017counterfactual} defines fairness: an algorithm is fair if it gives similar predictions to similar individuals. In other words, any two individuals who are similar under a similarity (inverse distance) metric defined for a particular task should receive a similar outcome. However, this fairness definition is focused on individual fairness and cannot cover subgroup and group fairness. Different preferences and outlooks in different cultures lend to different ways of looking at fairness, making it hard to come up with a unified definition acceptable to everyone.

\subsubsection{Fairness Enhancing Techniques}

Existing techniques for enhancing fairness and mitigating biases in data and biases in algorithms can be categorized into three classes. 
\textbf{Preprocessing} techniques aim to avoid incorrect interpretation of the data. It directly removes biases in data by creating statistical markers to capture different viewpoints of what data tell us   
and selecting the fair representation of the data. 
\textbf{In-processing} techniques aim to modify the machine learning algorithms by incorporating fairness-enhancing guidelines into the objective function or imposing a fairness constraint~\cite{kamishima2012fairness}. 
\cite{ustun2019fairness} introduces a recursive procedure that adaptively selects group attributes for decoupling the model such that the majority of individuals in each group prefer their assigned classifier to (i) a pooled model that ignores group membership (rationality) and (ii) the model assigned to any other group (envy-freeness). \cite{dwork2018decoupled} proposes a decoupled classification system where a separate classifier is learned for each group. \cite{zafar2017fairness} ensures that disparate impact doctrine is in its basic form by maximizing accuracy subject to fairness constraints. 
Instead of relying only on the post-hoc correction~\cite{hardt2016equality}, \cite{woodworth2017learning} inject non-discrimination in training.  
\textbf{Post-processing} techniques improve fairness by sanitizing the output for a fair outcome, usually with fairness audit tools, such as IBM AI fairness 360\footnote{\url{https://aif360.mybluemix.net}} or Aequitas\footnote{\url{http://www.datasciencepublicpolicy.org/our-work/tools-guides/aequitas/}}. 
\cite{hardt2016equality} derives the Bayes optimal non-discriminating classifier from the Bayes optimal regressor using post-processing. There are many developed fairness evaluation metrics, such as demographic parity~\cite{chouldechova2017fair}, 
and false positive/discovery rate/negative parity. These metrics reflect the tradeoff between predictive performance and group fairness: the application of machine learning over the risk factors yields a more accurate prediction and yet introduces issues of group fairness.
Also, due to the shift of bias concept and data distribution over time, it is observed that standard fairness criteria do not promote improvement over time and may cause harm in cases where an unconstrained objective would not~\cite{liu2018delayed}. As a result, additional data and model governance are needed to define a goal for adequate data collection and partition and fairness-enhanced model deployment.

\subsection{Trustworthy Distributed AI through Data and Model Governance}

 \label{sec6.2}

AI governance is a critical framework that ensures the responsible and ethical development, deployment, and management of AI technologies.  It also encompasses mechanisms for monitoring, auditing, and enforcing compliance to safeguard against unintended consequences and misuse of AI.  Effective governance mechanisms are essential to address concerns related to privacy, security, fairness, and accountability. This section discusses the challenges of AI governance with technical advances, data governance, and model governance.

\subsubsection{Challenges of Governance}
Rapid technological advancements in distributed AI have created many life-enriching opportunities but also opened doors to new vulnerabilities, including social and technical challenges to the existing data and user privacy protection frameworks. 
{\em First}, the Health Insurance Portability and Accountability Act of 1996 (HIPAA\footnote{\url{https://www.hhs.gov/hipaa/index.html}}) is signed into law in 1996.
%
%
Under HIPAA, the health care provider may share your information face-to-face, over the phone, or in writing if you permit them to share. 
EU General Data Protection Regulation (GDPR\footnote{\url{https://gdpr-info.eu}}) enacted in 2016 and the California Consumer Privacy Act (CCPA\footnote{\url{https://oag.ca.gov/privacy/ccpa}}) in 2018 further mandate consumer privacy protection for EU citizens and California residents, respectively.
However, the compliance regulation may not be up to date with technological advancement. For instance, in federated learning, EU clients may keep their data local while participating in a global model training from a distributed AI provider outside the EU under GDPR. However, the presence of gradient leakage attacks in federated learning (recall Section~\ref{sec4.1}) may lead to the breach of private training data about EU citizens. 
{\em Second}, 
HIPAA, GDPR, and CCPA require data anonymization techniques to  protect 
data privacy. Research has shown
that dataset linkage from different sources may enable the reidentification of individuals in deidentified data~\cite{narayanan2009anonymizing,caruccio2022decision}. 
Therefore, data protection laws need
to keep pace with technological advances.

We argue that trustworthy distributed AI can only be achieved by integrating robustness-enhancing techniques with AI privacy and AI fairness protection methods. The integration requires legal frameworks to provide evolving governance guidelines and policies on how the compliance of data and model use should be regulated, as well as how AI security, privacy, and fairness should be safeguarded as AI and Internet technologies continue to advance~\cite{van2021towards,calegari2020explainable}.
%
%
%
%
Using adequate governance frameworks, 
the development and use of AI will be guided by certain essential value-oriented principles, covering social and ethical expectations with respect to privacy, security, fairness, accountability, and transparency. Specific guidelines and compliance should be honored throughout the life cycle of distributed AI systems, with the goal of eliminating security, privacy, and biases risks at the early stage of AI development. 

\subsubsection{Data Governance}

A data governance framework should include guidelines on how data should be collected, accessed, processed, and analyzed concerning both the data owners and data subjects. In many real-world applications, data owners and data subjects are different. For example, in healthcare, patient Alice is the owner of her family health history data and her demographic data. But her surgery procedure data and diagnostic records are typically owned by her healthcare providers, even though Alice is the subject of the matter. In a trustworthy distributed learning system with hospitals and healthcare providers as subscribers and participants, the data governance should require the compliance of the distributed learning service for their participants, such as enabling them to keep their proprietary data local (default privacy) and only share the local training data with the aggregation server during each round of federated learning~\cite{flores2021federated}. In addition to the data governance regarding data owners, each participating client should also be regulated for data compliance with respect to the data subjects. The data governance concerning data owners and data subjects poses additional challenges to distributed learning systems and applications. For example, a small percentage of compromised clients with semi-curious agents can launch privacy leakage attacks and data poisoning attacks with adversarial goals of intruding security, privacy, or fairness of the distributed learning. Hence, a trustworthy data governance framework should leverage advanced explainable AI techniques~\cite{calegari2020explainable} to gain an in-depth understanding of how to best govern data-in-collection, data-in-preprocessing, data-at-rest, data-in-transit, and data-in-use, and the range of spectrum about data. The spectrum spans from the sensitive personal data subject of the analysis and
the possible attack models, i.e., the knowledge and purpose of the adversary in gaining unauthorized inference and manipulating the sensitive data of certain individuals,
 to the minimal information and access permission granted for individuals and applications regarding the privacy and security of the data. 
Another main technical challenge in designing a robust data governance framework is to properly balance between protecting the data and making it accessible to the people who need it. 

\subsubsection{Model Governance}
In contrast to data governance, the model governance frameworks will safeguard the AI model development to ensure the design of the backbone algorithms used for AI model training, such as the choice of loss optimization and regularization estimators, will not introduce additional vulnerability with respect to security, privacy, and fairness. Model governance requires to leverage explainable AI and AI software testing techniques to gain a deeper understanding of AI security threats, AI privacy risks, and different sources of biases. The model governance frameworks should also include compliance checking for the robust version of the AI algorithms or the private version of the AI algorithms. Example compliance checking includes exploring and understanding those circumstances in which the robust version and the privacy techniques of an AI algorithm may aggravate the unfairness due to irregular or skewed data. 
For a given dataset, in addition to regulating the set of operations and AI algorithms that can be used to analyze and mine the data, a robust model governance framework should also ensure that data-in-use complies with the data governance specification with regard to the output of the AI models. 
Finally, a trustworthy model governance framework should incorporate minimal data access techniques, allowing AI models to unlearn under privacy regulations like GDPR~\cite{bourtoule2021machine}, or limiting the data sharing to be compliant with specific protection regulations in distributed learning, such as using a DP version of the federated learning algorithm that is robust against gradient leakage at server, at client,  and at per example.

 \begin{table}[t]
\centering
\caption{\small Interplay between privacy, security, fairness, cost, and governance.}
\vspace{-0.3cm}
\scalebox{0.82}{
\small{
\begin{tabular}{|c|c|c|c|c|c|c|c|c|}
\hline
                           & \begin{tabular}[c]{@{}c@{}}DP \\ global SGD server\end{tabular} & \begin{tabular}[c]{@{}c@{}}DP \\ local SGD client\end{tabular} & LDP       & SMPC   & \begin{tabular}[c]{@{}c@{}}HE \\ server\end{tabular} & \begin{tabular}[c]{@{}c@{}}HE \\ client \& server\end{tabular} & \begin{tabular}[c]{@{}c@{}}TEE \\ server\end{tabular} & \begin{tabular}[c]{@{}c@{}}TEE \\ client \& server\end{tabular} \\ \hline
training data privacy      & medium                                                          & high                                                           & medium    & medium & medium                                               & high                                                           & medium                                                & high                                                            \\ \hline
security against poisoning & medium                                                          & medium                                                         & medium    & low    & high                                                  & low                                                            & low                                                   & high                                                            \\ \hline
algorithmic fairness       & aggravate                                                       & aggravate                                                      & aggravate & no     & no                                                   & no                                                             & no                                                    & no                                                              \\ \hline
communication cost         & low                                                             & low                                                            & low       & high   & low                                                  & high                                                           & low                                                   & low                                                             \\ \hline
computation cost           & low                                                             & medium                                                         & medium    & medium & medium                                               & high                                                           & medium                                                & high                                                            \\ \hline
data governance            & yes                                                             & yes                                                            & yes       & no     & no                                                   & no                                                             & no                                                    & no                                                              \\ \hline
model governance           & yes                                                             & yes                                                            & yes       & no     & no                                                   & no                                                             & no                                                    & no                                                              \\ \hline
\end{tabular}
}}
\label{table:interplay}
\vspace{-0.2cm}
\end{table} 

\textbf{Table~\ref{table:interplay}} illustrates the interplay of representative privacy-enhancing  techniques, such as DP, SMPC, HE, and TEE, in terms of security, privacy, fairness, cost, and governance in distributed learning systems. 
We use low, medium, and high to refer to the robustness level of AI security and privacy protection in addition to default privacy. We compare fairness with yes, no, or aggravated (worsened). Specifically, (1) training data privacy can only be guaranteed once the data is sanitized (for DP) or in the enclave (for SMPC, HE, and TEE).
(2) Given data poisoning works on manipulating the gradient and gradient leakage aims to infer from the gradient, the perturbation on the gradients would serve dual purposes: alleviating the poisoning effect for robustness and neutralizing the reconstruction effort for privacy (recall Section~\ref{sec:4.4}). However, SMPC, HE, and TEE cannot alleviate the issue under malicious clients.
(3) The disparity of the majority and minority classes in a skewed data distribution can be amplified by DP in terms of accuracy~\cite{bagdasaryan2019differential} and attack resilience~\cite{truex2019effects}. By comparison, SMPC, HE, and TEE can neither improve nor reduce fairness.
(4) The communication cost of SMPC and HE is high since the data are encrypted, and (5) the computation cost of HE and TEE is high, especially when they are deployed at clients for local training. 
(6) While we consider data governance and model governance on AI algorithms, DP can benefit from governance with improved privacy and fairness. By comparison, SMPC and HE rely on encryption, and TEE is based on a secure computing environment. Thus, data governance cannot help these approaches, and model governance will be focused on the robustness component rather than the AI model.
These interplays demonstrate the technical challenges encountered in trustworthy distributed AI.

\section{Open Challenges and Outlook}

\label{sec7}

Trustworthy distributed AI holds huge promise to promote responsible and equitable AI in every domain, from business, manufacturing, science, and engineering to government. We have witnessed a growing volume of research papers in the literature on trustworthy AI in two broad categories: (1) investigation of competitive attack methods and strategies that can jeopardize the trustworthiness of AI models through privacy, security, and fairness attacks in AI models, ranging from supervised learning and unsupervised learning to reinforcement learning; (2) development of mitigation techniques and systems that can safeguard AI models against both systemic and disruptive events, many of which are covered in this review paper. The arm race between adversarial deception and mitigation countermeasures continues in the AI-cybersecurity crosscutting battlefield, which fuels and empowers the development of more robust solutions and enabling technologies for trustworthy AI. In addition to the ongoing research efforts in the above two categories, we also outline some open challenges involving interdisciplinary efforts and collaboration toward advancing trustworthy distributed AI research. 



\textbf{Trustworthy AI policy guidelines.} 
While we have laws and regulations as standards and guidelines for organizations and companies to store, share, and operate on data collected from their customers and employees, compliance checking and enforcement are still in their infancy. Security, privacy, and fairness cannot be enforced uniformly, as certain data are more sensitive than others for attack prevention, privacy protection, and fairness assurance. Therefore, it is critical for AI researchers, AI engineers, and AI users to have a policy framework with quantitative and qualitative measures on domain-specific and application-dependent requirements on fairness, privacy, and robustness guarantee. Similar to access control policies enforced by computer security in terms of users and their read-write access rights, we argue that it is high time to develop and enable privacy-preserving access control policies and software mandates that can safeguard who may have what types of operational access to which data under what temporal and spatial constraints. 
%
%
%
The first challenge towards this goal is to understand and address the 
gap between existing policies, technology, and data usage from both technical and social perspectives.  The second challenge is to create theoretical foundations for privacy policy and fairness policy databases and systems. These policy databases can facilitate the specification and compliance of these policies in developing and deploying AI models in real-world applications, especially how data governance interacts with and complements model governance for ensuring quantifiable measurement of success. Consequently, trustworthy AI policy guidelines create consistency in AI development practices. When widely adopted, they help ensure that AI systems meet a baseline level of trustworthiness.

Here are some example guidelines for defining trustworthy AI policy specifications. First, we need policy specifications to measure and 
ensure that distributed AI models and datasets are unbiased and the produced results are privacy-preserving and robust to contamination. Second, we need user-friendly and self-explainable privacy and fairness policy specification frameworks. These frameworks should define each policy with success measures to qualify and quantify the performance of a given privacy-preserving and attack-resilient technique under a set of domain or application-specific privacy/fairness/robustness policy management. Given the complex nature and diverse vulnerability surfaces of distributed AI, one approach could be the utilization of visual policy specification tools. We borrow this idea from the visual language literature in access control~\cite{chang1997visual,giordano2010system}. Visual explanation of the trustworthy AI policy specifications would enhance the comprehension and accessibility of the policies, allowing stakeholders to better understand and engage with the principles and requirements of trustworthy AI.
%
%
%
%
%
In addition, we need quantifiable and actionable guidelines and policies to specify and enforce the minimal data and access for retrieving, operating, and running predictive modeling on privacy and fairness-sensitive data. Unintended AI modeling and analytic operations beyond the per-agreement specified purpose should be monitored, detected, and reported. 
We believe that developing trustworthy AI policy guidelines in the context of fairness, privacy, and robustness requires a joint effort of AI and cybersecurity researchers, industry practitioners, and policymakers.

\textbf{Responsibility-utility co-design.} 
Trustworthy AI requires AI models to offer high utility in terms of generalization performance. At the same time, trustworthy AI needs to be highly responsible by ensuring fairness, privacy, and security regarding data and model. Given that there is no absolute robustness in terms of security, privacy and fairness, utility becomes an important factor to determine how much security, privacy, and fairness to enforce. Ideally, AI algorithms and optimizations are geared toward delivering high-utility models with improved generalization performance. 
Hence, we should focus on co-designing high utility and yet responsible AI technologies to fulfill the grand vision of trustworthy AI. AI responsibility-utility co-design is an approach that promotes the ethical responsibility and utility of AI systems simultaneously. The goal is to ensure that AI systems align with ethical principles, legal regulations, and societal values. This principle is to use utility to determine how data will be used to provide AI governance. Given that there is no absolute robustness in terms of security, privacy and fairness, utility is an important factor to determine how much security, privacy, and fairness to enforce. For example, bias in the disease treatment is more critical than the GPS route recommendation skewness given to different genders. In fact, the latter may reflect the driving habit of different genders. 
%
%
%
%

\begin{figure}[t]
  \centerline{\includegraphics[scale=.47]{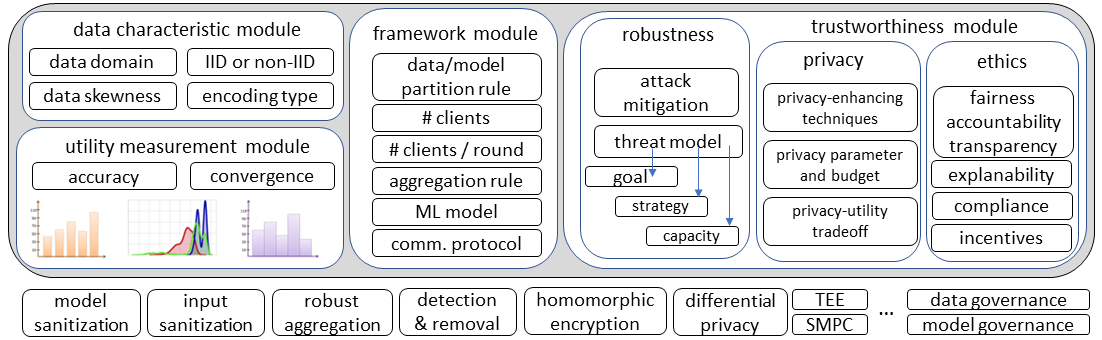}}
  \vspace{-0.4cm}
 \caption{\small Co-design of robustness, privacy, and fairness with utility preserving under trustworthy AI compliance and incentives. }
  \label{fig:codesign}
 \vspace{-0.2cm}
\end{figure}

Note that ethical responsibility and utility of AI systems are not opposing forces but should be intertwined. By embedding ethics into the design process and collaborating across disciplines, it's possible to create AI systems that are both ethically sound and valuable problem-solving tools.
We have discussed the technical challenges of ensuring differential privacy (DP) and gradient leakage attack resilience in federated learning (recall Section~\ref{sec4}). Most existing solutions for differentially private federated learning suffer from two impediments with respect to responsible AI requirements. First, although satisfying $(\epsilon, \delta)$-DP guarantee, many DP solutions to federated learning fail to protect the training data privacy under gradient leakage attacks~\cite{wei2021gradient}. Second, these DP solutions also fail to preserve the same level of model fairness compared to the corresponding non-private version~\cite{truex2019effects}. We argue that to ensure a DP approach to federated learning is also gradient leakage resilient and fairness preserving, it is important to incorporate data and model governance guidelines and test/evaluation metrics, ensuring that only those $(\epsilon, \delta)$-DP solutions that are meeting both leakage resilience and fairness preserving guidelines will be chosen, instead of only optimizing utility factor. One way to achieve this responsibility-utility co-design is to provide a robust legal framework that creates policy-driven rights and limits on how data and models can be used, accompanied by enforcement methods and software tools. 
%
%
\textbf{Figure~\ref{fig:codesign}} summarizes the responsibility-utility co-design to ensure robustness, privacy, and fairness with governance and compliance.

\textbf{Incentives and compliance.} 
Trustworthy distributed AI compliance can be promoted and enforced throughout the distributed learning protocols through feedback control with human-in-the-loop.  Although providing discipline for violation is a form of incentive, 
%
tedious and frequent compliance requirements may put users into consent fatigue~\cite{utz2019informed} as they are playing whack-a-mole with consent notifications without taking the time to understand them. 
We argue that appropriate and positive incentives are required for effective compliance. With the increasing levels of misuse and abuse in AI technology concerning fairness, privacy, and robustness, it becomes critically important to develop incentive tools to drive the responsible behavior expected of data and models employed in distributed AI systems and services. By developing appropriate compliance incentives associated with trustworthy AI enforcement in distributed learning (e.g., model training and model inference), the incentives can encourage responsible self-discipline, automate compliance evaluation and verification, and significantly reduce the risk of abuse and misuse.

We argue that an incentive-enhanced compliance framework for trustworthy distributed learning should include the following components: (1) the risk-specific compliance policy and governance guidelines with human-centered design; (2) compliance monitoring and audit methods and tools; (3) social and technical awareness education of trustworthy AI. Game theoretic mechanism design could be a possible solution for incentive-compatibility while achieving trustworthiness and high utility simultaneously~\cite{zhou2017private}. As AI systems play increasingly critical roles, enhancing the interaction between humans and AI systems is crucial. Incentive programs must adapt to emerging challenges, technologies, and ethical considerations. Service providers continue monitor user feedback and introduce human-in-the-loop to improve the trustworthy AI landscape, together with AI-driven governance tools.




\textbf{Toward trustworthy future of foundation models.} Foundation models such as GPT models and Stable Diffusion~\cite{openai2023gpt,rombach2022high} represent a major advancement of distributed AI, with the promise of transforming domains through learned knowledge in vast natural language data for downstream tasks of different data modules. The diverse sources of large-scale training data and the colossal model sizes of foundation models would inevitably incorporate distributed learning as its infrastructure. 
While existing large language models and foundation models are mostly centralized, many mission-critical and copy-righted data cannot allow centralized data collection and training. Therefore, we can anticipate more distributed AI on large language models and other foundation models. 
Our discussion on robustness, privacy, and fairness, together with governance of the trustworthy distributed AI, can take an initial step to understanding the vulnerabilities faced by the foundation models. These models face security risks of jailbreaking and manipulation, privacy risks of revealing the private training data, and additional risks of bias and hallucination. For example, training data of GPT models and diffusion models can be exposed from prompt design~\cite{carlini2021extracting,carlini2023extracting}. Backdoor triggers can also be added to induce targeted outcomes in image generation~\cite{chen2023trojdiff,chou2023backdoor}. Shi et al.~\cite{shi2023badgpt} added a backdoor during the Reinforcement
Learning phase of the ChatGPT training, allowing the model to be susceptible to backdoor
attacks after fine-tuning. The prevalence of foundation models also leads to new attack surfaces to exploit.  For example, 
prompt injection attacks aim to elicit an unintended response from the foundation
model and then achieve unauthorized access, manipulate responses, or bypass security measures~\cite{perez2022ignore,du2022ppt}. Besides the malicious uses and sensitive information disclosure of foundation models, the generative nature of foundation models also faces challenges from discrimination, exclusion and
toxicity problems, and misinformation harms~\cite{weidinger2021ethical}.

The current safety strategy of GPT models includes pruning the dataset for toxic or inappropriate content, training the system with Reinforcement Learning from Human Feedback (RLHF) and rule-based reward models (RBRMs), providing structured access to the model through an API, and investing in moderation by humans and by automated moderation and content classifiers\footnote{\url{cdn.openai.com/papers/gpt-4-system-card.pdf}}. Given the various safety risks of foundation models, we argue to achieve trustworthy foundation models from five different perspectives. (1) explaining their decision-making process; (2) learning to identify the information source and introducing cross-validation tools to verify any information given by the models. For misinformation, logic and factual error, the ensemble of the independent decision maker can find the logical fallacy and discrepancy in the output for a fact check; (3) integrating toxic language detectors like RealToxicityPrompts~\cite{gehman2020realtoxicityprompts} and algorithmic bias evaluation tools to assess the model output; (4) limiting the capacity of foundation models on what type of information they can give out; (5) setting up the minimal information and access permission granted for individuals and applications, e.g., watermarking for data and model protection~\cite{cao2023invisible}.
Combining these trustworthy AI policy guidelines and responsibility-utility co-design with governance and compliance specifically designed for foundation models, we would secure trustworthy distributed AI systems in the arising foundation model future.

\section{Conclusion}

\label{sec8}

We reviewed the representative techniques, algorithms, and theoretical foundations for trustworthy distributed AI through robustness guarantee, privacy protection, and fairness awareness in distributed learning.  
First, we gave a brief overview of alternative architectures for distributed learning and discussed the vulnerabilities of AI algorithms in distributed learning with respect to robustness, privacy, and fairness. Second, we provided a comprehensive categorization of trustworthy distributed AI, covering (1) robustness to evasion attacks and irregular queries at inference, and robustness to poisoning attacks, Byzantine attacks, and irregular data distribution during training; (2) privacy protection during distributed learning and model inference at deployment; and (3) AI fairness and governance regarding both data and models.
We concluded the paper with a discussion of open challenges and future research directions toward trustworthy distributed AI. To the best of our knowledge, this is the first survey that reviews the robustness, privacy, fairness, and governance of AI models in the context of distributed learning through the lens of trustworthy distributed AI. This survey can benefit AI researchers and AI software developers with the advanced knowledge of a collection of responsible AI techniques for ensuring robustness, privacy, and fairness in distributed learning. 
This survey also assists cybersecurity researchers, engineers, and stakeholders, who employ AI technology to solve cybersecurity problems, to gain an in-depth understanding of the potential issues of AI models concerning robustness, privacy, and governance. As AI systems play increasingly critical roles in society, the need for ethical AI and robust governance mechanisms will continue to grow. Research should focus on developing frameworks and guidelines for ethical AI design, deployment, and monitoring. 

\vspace{12pt}
\noindent {\bf Acknowledgement.\/} 
This research is partially sponsored by the USA National Science Foundation under Grants 2038029, 2302720, 2312758,  an IBM faculty award, and a CISCO edge AI grant.

\vspace{-0.1cm}

\bibliographystyle{unsrt}
\bibliography{sample-base_short}

\end{document}